%% file: main.tex
\documentclass{article}

\usepackage[preprint]{neurips_2026}

\usepackage[utf8]{inputenc} 
\usepackage[T1]{fontenc}    
\usepackage{hyperref}       
\usepackage{url}            
\usepackage{booktabs}       
\usepackage{amsfonts}       
\usepackage{nicefrac}       
\usepackage{microtype}      
\usepackage{xcolor}         

\usepackage{graphicx}
\usepackage{float}
\usepackage{amsmath}
\usepackage{algorithm}
\usepackage{algorithmic}
\usepackage{wrapfig} 
\robustify{\hyperlink}
\robustify{\hypertarget}

\title{GauS: Differentiable Scheduling Optimization via Gaussian Reparameterization}

\author{%
    Yaohui Cai \\
    Cornell University \\
  \texttt{yc2632@cornell.edu} \\
  \And
      Vesal Bakhtazad \\
    Cornell University \\
  \texttt{vb354@cornell.edu} \\
  \AND
      Cunxi Yu \\
    University of Maryland, College Park \\
  \texttt{cunxiyu@umd.edu} \\
  \And
      Zhiru Zhang \\
    Cornell University \\
  \texttt{zhiruz@cornell.edu} \\
}

\newcommand{\new}[1]{{\color{black}#1}}

\begin{document}

\maketitle
\input{0-abstract}

\input{1-introduction}
\input{2-preliminary}
\input{3-method}

\input{4-evaluation}
\input{5-conclusion}





\clearpage
\bibliographystyle{plainnat}
\bibliography{references}

\clearpage
\newpage
\input{a1-appendix}
\input{checklist.tex}

\end{document}

%% file: 0-abstract.tex
\begin{abstract}
Efficient operator scheduling is a fundamental challenge in software compilation and hardware synthesis.
While recent differentiable approaches have sought to replace traditional ones like exact solvers or heuristics with gradient-based search,
they typically rely on categorical distributions that fail to capture the ordinal nature of time and suffer from a parameter space that scales poorly.
In this paper, we propose a novel differentiable framework, GauS, that models operator scheduling as a stochastic relaxation using Gaussian distributions, which fully utilize modern parallel computing devices like GPUs.
By representing schedules as continuous Gaussian variables, we successfully capture the ordinal nature of time and reduce the optimization space by orders of magnitude.
Our method is highly flexible to represent various objectives and constraints, which provides the first differentiable formulation for the complex pipelined scheduling problem.
We evaluate our method on a range of benchmarks, demonstrating that GauS achieves Pareto-optimal results.
\end{abstract}

%% file: 1-introduction.tex
\section{Introduction}
The efficient scheduling of operations onto a finite set of resources,
an NP-hard problem known as operator scheduling,
is a fundamental challenge across the entire computing stack.
In software compilation, it is the critical phase that determines the execution order of instructions to maximize instruction-level parallelism and minimize pipeline stalls~\cite{lam1988software}.
In hardware synthesis, it is the cornerstone of high-level synthesis (HLS)~\cite{cong2006efficient, ku2002relative, radivojevic1996new}, where it dictates the cycle-by-cycle behavior of custom logic and the allocation of physical functional units.
At its core, the problem involves mapping a set of operations to a discrete set of time steps while minimizing a given cost function and honoring a set of constraints. 

As we transition into an era of massive-scale deep learning models and heterogeneous architectures, the complexity of scheduling tasks has scaled dramatically from hundreds to tens of thousands of nodes.
Modern workloads, such as those targeting the Groq Tensor Streaming Processor (TSP)~\cite{abts2020think},
require orchestrating thousands of functional units simultaneously to exploit massive data parallelism.
Similarly, emerging Processing-in-Memory (PIM) architectures~\cite{guo2025pimsynth, seshadri2017ambit}, 
present unique scheduling challenges where bit-serial execution across large memory arrays requires precise temporal and spatial specification of thousands of operations.

Historically, efforts to solve the operator scheduling problem have fallen into two disparate camps, neither of which fully satisfies the needs of modern large-scale systems~\cite{leiserson1991retiming}.
On one hand, Integer Linear Programming (ILP)~\cite{fan2005cost} and Satisfiability Modulo Theory
(SMT)~\cite{fan2008modulo} solvers provide optimal solutions but suffer from exponential runtime complexity, making them impractical for realistic workloads~\cite{amaru2015epfl, soi2025optimal}.
On the other hand, heuristic-based methods, such as list scheduling~\cite{parker1986maha, kondratyev2011realistic} and force-directed scheduling (FDS)~\cite{verhaegh1992force, paulin1987force},
offer the speed required for large-scale graphs but often yield suboptimal results,
as they lack a global view of the optimization space and struggle to balance competing constraints like memory footprint and communication overhead.

A promising third path has recently emerged: differentiable combinatorial optimization.
which has been widely used in many real-world combinatorial problems ~\cite{bengio2021machine, zhang2025cypress, cai2025smoothe,lin2019dreamplace}.
By relaxing discrete scheduling decisions into a continuous space,
these methods transform the NP-hard combinatorial search into a smooth, differentiable landscape.
This shift provides a holistic view of the global optimization problem and allows the problem to be solved by gradient descent and utilizing the modern ML ecosystem including parallel computing devices like GPUs. 
While GS-Schedule~\cite{liu2024differentiable} as an early attempt in this domain successfully proved the feasibility of differentiable optimization for scheduling using categorical relaxations (i.e., Gumbel-Softmax),
it does not fully utilize the parallelism of GPUs and the number of parameters requiring optimization increases linearly with respect to the depth ($D$) and size ($|V|$) of the graph. 

In this paper, we introduce GauS, a differentiable \underline{Gau}ssian reparameterization \underline{S}cheduling framework.
We model the execution step of each operator as a continuous random variable following a Gaussian distribution $X\sim \mathcal{N}(\mu, \sigma^2)$,
where the mean ($\mu$) represents the expected schedule step and the standard deviation ($\sigma$) represents the uncertainty.
Under this formulation, the number of parameters requiring optimization reduces to $2|V|$.
The probability density function (PDF) of random variables provides a smooth and continuous proxy for discrete placement.
Because of the properties of Gaussian distribution, we can represent a wide spectrum of expectations of objectives and constraint violations as differentiable functions of the Gaussian parameters ($\mu$ and $\sigma$).
By minimizing the weighted sum of them,
the original discrete combinatorial scheduling problem is essentially transformed to a continuous problem that can be efficiently solved by gradient descent.

\new{Our contributions are: \textbf{(i) Novel Representation}: the first use of Gaussian reparameterization in operator scheduling, which naturally encodes temporal proximity and reduces the parameter space from $O(D|V|)$ to $O(|V|)$; \textbf{(ii) Differentiable Flexibility}: differentiable expressions for diverse objectives and constraints, enabling the first differentiable formulation of pipelined scheduling~\cite{lam1988software, rau1981some}.}



Through extensive empirical evaluation on both synthetic and realistic benchmarks, we demonstrate that our method consistently produces the Pareto-optimal frontier of speed and quality. 
Compared to previous differentiable method and commercial solvers, GauS achieves equivalent solution quality while one to two orders of magnitude faster

%% file: 2-preliminary.tex
\section{Preliminaries}
\subsection{Scheduling Problem}
Scheduling is the process of assigning a set of operations to specific time steps, while satisfying given constraints and optimizing for performance objectives. 
In this section, we formalize the objectives and constraints for both regular and pipelined scheduling.

\subsubsection{Regular Scheduling}
A workload is represented as a directed acyclic graph (DAG), $G = (V, E)$,
where vertices $V = \{v_1, \dots, v_n\}$ represent computational operations and edges $E$ denote data dependencies.
In a regular scheduling problem, each operator $v_i$ is assigned a discrete start time step $s_i \in \mathbb{N}$

\textbf{Dependency constraint (\underline{Dep})}.
The consumer $v_j$ cannot begin until the producer completes its execution:
\begin{equation}
\text{\underline{Dep}:}\quad \forall v_i, v_j \in E, \quad s_i + \text{Lat}(v_i) \le s_j
\end{equation}
where Lat($v_i$) is the execution latency of $v_i$.
In contexts allowing \textit{operation chaining}, 
where multiple operations with zero latency can be scheduled in the same step.
Then this simplifies to $s_i \le s_j$.
For the sake of simplicity, we assume Lat$(v_i)=1, \forall i$ in the remaining discussion.

\textbf{Latency constraint (\underline{Lat})} ensures the schedule length does not exceed a predefined limit $D$:
\begin{equation}
\text{\underline{Lat}:}\quad \max s_i \le D
\end{equation}

\textbf{Resource usage}.
Let $r_i$ denote the resource demand of operator $v_i$.  
The resource usage at step $d$, $Res(d)$ is the sum of resources required for all operators active at step $d$, and the global resource usage is the maximum resource usage over all steps:
\begin{equation}
    Res(d) = \sum_{v_i \in V} r_i \cdot \mathbb{I}[d_i=d], \quad
    \mathcal{L}_{Res} = \max_{d}  Res(d)
\end{equation}
Here $\mathbb{I}[C]$ is the indicator function, which outputs $1$ when condition $C$ is met and $0$ otherwise.

\input{figures/schedule}

\textbf{Memory Footprint}.
To carry data across steps, operator $v_i$ utilizes $b_i$ storage units to store its output.
Depending on the problem settings, storage units could be registers, SRAM, or DRAM.
A storage units is occupied from the step $s_i$ until the step of its latest successor starts.
Let $succ(v_i) = \{v_j \mid (v_i, v_j) \in E\}$ denote the set of immediate successors of $v_i$.
The memory usage at step $d$ is the sum of all active storage units which is required by all operators $v_i$ scheduled at or before step $d$ whose outputs are used by at least one operator scheduled after step $d$.
And the global memory footprint is the maximum memory usage over all step\footnote{We add an explicit sink node as a successor for all leaf nodes (nodes without successors) to ensure their outputs are also carried toward the end.}:
\begin{equation}
    Mem(d) = \sum_{v_i \in V}  b_i \cdot \mathbb{I}\left[ s_i \le d < \max_{v_j \in succ(v_i)} s_j \right], \quad
    \mathcal{L}_{Mem} =  \max_d Mem(d)
    \label{eq:mem}
\end{equation}

Figure~\ref{fig:ill} demonstrates how different objectives impact the schedule of the same workload.
A resource usage optimized schedule is on the left and a storage unit optimized schedule is on the right.

\subsubsection{Modulo Scheduling}
For high-throughput applications,
\textit{modulo scheduling}~\cite{rau1981some, lam1988software} achieves pipelined execution by starting a new loop iteration (or function invocation) every \textit{initiation interval} (II) steps.
Because schedules from consecutive iterations overlap, the lifetimes of an operator and its output are wrapped within the initiation interval, which affects the resource requirements.
Consequently, an operator scheduled at any step $d$ such that $d = t+ k\cdot \text{II}, k\ge 0$, will demand resources ($MRes$) and occupy storage units ($MMem$) at time step $t, 0\le t < \text{II}$:
\begin{equation}
    MRes(t) =  \sum_{k} Res(t + k\cdot \text{II}), \quad
    MMem(t) =  \sum_{k} Mem(t + k\cdot \text{II})
\end{equation}


\textbf{Recurrence Constraint (\underline{Rec}):}
With loop-carried dependencies between iterations, a back-edge from $v_j$ to $v_i$ with a distance $k$ (iterations) represents the output from $v_j$ is required for a $v_i$ after $k$ iterations.
This constraint effectively sets a deadline for operator $v_j$ relatively to the operator $v_i$.
\begin{equation}
    \text{\underline{Rec}:} \quad s_i + k \cdot \text{II} \ge s_j + \text{Lat}(v_j)
\end{equation}

\subsection{Scheduling Algorithms}

\new{
\textbf{Exact approaches} formulate scheduling as ILP~\cite{hwang2002formal, gebots1993cost, fan2005cost} or SAT~\cite{fan2008modulo, zhang2004sat}; they guarantee optimality but scale exponentially and rarely converge on realistic graphs.

\textbf{Heuristics} such as ASAP/ALAP, List Scheduling~\cite{parker1986maha, kondratyev2011realistic}, and Force-Directed Scheduling (FDS)~\cite{verhaegh1992force, paulin1987force} trade quality for speed via greedy local decisions; for \textit{modulo scheduling}, where cyclic resource coupling and recurrence constraints often conflict with the target II, ad hoc greedy heuristics dominate~\cite{cong2006efficient, radivojevic1996new, ku2002relative}.

\textbf{Differentiable methods}~\cite{bengio2021machine, lin2019dreamplace} relax discrete decisions into continuous distributions, enabling first-order optimization, GPU vectorization, and smooth surrogates for non-linear objectives such as memory footprint or learned cost models~\cite{chen2024syn, wu2023gamora}.
GS-Schedule~\cite{liu2024differentiable} first applied this paradigm to scheduling, modeling each placement as a categorical distribution sampled via the Gumbel-Softmax trick~\cite{jang2016categorical}.
}

%% file: figures/schedule.tex
\begin{figure}
    \centering
    \includegraphics[width=0.4\linewidth]{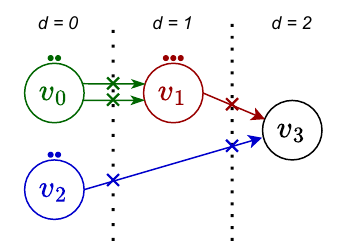}
    \includegraphics[width=0.4\linewidth]{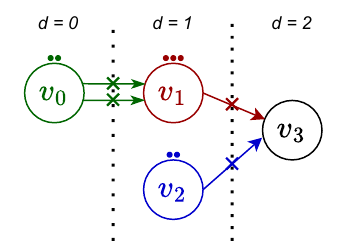}
    \caption{
    \textbf{Impact of optimization objectives on scheduling} ---
    \small{
    Impact of optimization objectives on scheduling.
    Nodes $v_i$ (operators) show resource requirements $r_i$ (number of dots) and bitwidths of storage units required $b_i$ (number of arrows).
    Max depth ($D$) sets to $3$.
    Left schedule minimizes resources ($\mathcal{L}_{res}=4, \mathcal{L}_{mem}=3$);
    right schedule minimizes memory footprint ($\mathcal{L}_{res}=5, \mathcal{L}_{mem}=2$).
    }}
    \label{fig:ill}
\end{figure}

%% file: 3-method.tex
\section{Method}
\subsection{Gaussian Reparameterization}
\label{sec:reparam}
The previous differentiable scheduling attempt~\cite{liu2024differentiable},
represents the placement of an operator $v_i$ as a categorical distribution over $D$ discrete steps,
which requires a parameter size of ${D\cdot|V|}$.
Such a formulation suffers from two primary limitations:

\textbf{Ordinal blindness}:
    Categorical distributions treat steps as independent, nominal labels.
    From the optimizer's perspective, the relationship between step $d$ and $d+1$ is mathematically identical to that between $d$ and $d+100$.
    This ignores the fundamental ordinal nature of time; 
    in scheduling, proximity in the timeline should imply proximity in the optimization space.
    
\input{figures/gaussian}
\textbf{Insufficient scalability}:
    the optimization space $\mathbb{R}^{D\cdot|V|}$ becomes prohibitively large ,
    as either maximum depth $D$ or the number of operators $|V|$ increases.
    This limitation becomes increasingly prominent as the needs for large-scale modern workloads scheduling~\cite{abts2020think, soi2025optimal} increases.
    

To address these limitations, we propose \textbf{Gaussian reparameterization}.
In stead of $D$ categorical probabilities, we model the schedule of each operator $v_i$ as an independent Gaussian random variable $X_i$: 
$X_i \sim \mathcal{N}(\mu_i, \sigma_i^2)$,
where $\mu_i \in \mathbb{R}$ represents the expected scheduling step and $\sigma_i \in \mathbb{R}^{+}$ represents the standard deviation of the distribution.
The probability density function (PDF), $p(x; \mu_i, \sigma_i)$, represents the likelihood of an operator being scheduled at any continuous point $x$ along the timeline.
Figure \ref{fig:gaussian} illustrates Gaussian reparameterization.
\new{Note that our framework is not limited to Gaussian. We empirically show that other \textit{smooth symmetric unimodal CDFs} can perform similar in \ref{app:dist}.}

By adopting this parameterization, we reduce the number of parameters to be optimized from $D\cdot |V|$ to $2|V|$.
For a graph with $N$ operators, we only need to optimize two vectors:
mean vector ($\boldsymbol{\mu} \in \mathbb{R}^{|V|}$) that determines the central temporal location of each operator, 
and standard deviation vector ($\boldsymbol{\sigma}\in {\mathbb{R}^+}^{|V|}$) that controls the degree of relaxation.
During initial stage of optimization,
$\boldsymbol{\sigma}$ is typically large to encourage encourages exploration of the schedule space;
as optimization processes, $\boldsymbol{\sigma}$ usually converges to small values concentrating the probability mass and
effectively hardening the schedule toward a deterministic discrete integer.

\new{This Gaussian relaxation also introduces a crucial \textbf{temporal proximity} prior. The categorical formulation of GS-Schedule~\cite{liu2024differentiable} represents each step as an orthogonal one-hot vector, making all stages mutually equidistant in optimization space even though adjacent steps are temporal neighbors in the schedule space; a small shift in $\boldsymbol{\mu}$ under our Gaussian PDF instead smoothly transports probability mass between neighboring steps, so the gradient pulls each operator toward its temporal neighbors rather than across a flat one-hot simplex.}
\new{We empirically isolate the temporal proximity effect in Appendix~\ref{app:temporal}: after removing the temporal proximity by a controlled $d$-axis shuffle, EPFL~\cite{amaru2015epfl} benchmarks degrade with an average of $35\%$.}

\subsection{Representing Objectives and Constraints}
\label{sec:drv}

With the Gaussian reparameterization established,
the scheduling optimization is transformed into minimizing the expectations of performance objectives and constraint violations with respect to the parameters ($\boldsymbol{\mu}, \boldsymbol{\sigma}$).
By optimizing the expected values, we ensure that the continuous parameter space accurately reflects the costs of the discrete schedules sampled from it.
We illustrate this through two important metrics used in regular scheduling;
the full derivations, including pipelined scheduling, are provided in Appendix~\ref{app:drv}.

As emphasized in~\cite{lin2019dreamplace, liu2024differentiable, cai2025smoothe}, 
It is important to maintain consistency between optimization and the final sampling process for differentiable combinatorial optimization. 
In our framework, to extract a discrete schedule from continuous variables, we round $X_i$ to the nearest integer.
Therefore, we use probability $P_i^d$ to represent operator $v_i$ being scheduled at step $d$ by integrating the Gaussian PDF over the unit interval centered at $d$, 
which is used to align the sampling and optimization:
\begin{equation}
P_i^d = P\left\{ \lfloor X_i + 0.5 \rfloor = d \right\}
= \Phi\left(\frac{d+0.5-\mu_i}{\sigma_i}\right) - \Phi\left(\frac{d-0.5-\mu_i}{\sigma_i}\right) 
\label{eq:round}
\end{equation}
\label{eq:def}
Here $\lfloor \cdot\rfloor$ denotes  truncation and $\Phi(\cdot)$ is the standard Gaussian cumulative distribution function (CDF). 
This formulation is efficient and differentiable.
Notably, the latency constraint (\underline{Lat}) is implicitly enforced by bounding the range of $d$ within $\{0, \dots, D-1\}$
\footnote{The lower limit for the first step ($d=0$) and the upper limit for the last step ($d=D-1$) are set to $-\infty$ and $+\infty$, respectively.}.
Given $P_i^d$, the expectation of global objectives and violations of constraints can be expressed as differentiable functions of $\boldsymbol{\mu}$ and $\boldsymbol{\sigma}$.

For \textit{dependency constraint}, 
we calculate the expected number of violations across all edges.
A violation occurs if 
a consumer $v_j$ is scheduled at a step $d_j$ that is earlier than the producer $v_i$'s step $d_i$.
Thus the expected violations of dependency constraint $\widehat{\mathcal{V}}_{Dep}$ is
\begin{equation}
    \mathcal{\widehat{V}}_{Dep} = \mathbb{E}[\mathcal{V}_{Dep}] 
    = \sum_{(v_i, v_j) \in E} \sum_{d_i=1}^{D-1} \sum_{d_j=0}^{d_i - 1} P_i^{d_i} \cdot P_j^{d_j} 
\label{eq:dep}
\end{equation}

\textit{Memory footprint} requires calculating the probability that an operator's output is active at step $d$. 
A storage unit for $v_i$ is occupied if the operator has started ($X_i \le d$) but at least one successor $v_j$ has not yet been scheduled ($X_j > d$). 
The expected storage unit count at step $d$ is:
\begin{subequations}
\begin{align}
\widehat{Reg}(d) = \mathbb{E}[Reg(d)] &= \sum_{v_i\in V} P\left\{ X_i \le d < \max_{v_j \in succ(v_i)} X_j \right\} \cdot b_i  \label{eq:active} \\
&= \sum_{v_i\in V} \left( \sum_{d_i=0}^{d} P_i^{d_i} \right) \cdot \left( 1 - \prod_{v_j \in succ(v_i)} \sum_{d_j=0}^{d} P_j^{d_j} \right) \cdot b_i
\end{align}
\end{subequations}
The global memory footprint is defined by the max pressure across all steps.
Since the \texttt{max} function is non-differentiable,
we employ the \texttt{LogSumExp} operator as a smooth, differentiable approximation for the expectation of \texttt{max}, where temperature parameter $\tau$ controls how closely \texttt{LogSumExp} approximates \texttt{max}.
\begin{equation}
\widehat{\mathcal{L}}_{Reg} = \mathbb{E}\left[\max_s Reg (d)\right]
\approx \tau\cdot \log \left(\sum_s \exp \left(\frac{\widehat{Reg}(d)}{\tau}\right) \right)
\label{eq:max}
\end{equation}

\subsection{Overall Algorithm}
\textbf{Optimization}.

\new{The expected objectives and violations turn scheduling into a continuous constrained optimization problem.
Rather than tuning static penalty weights $\lambda_i$ for $\min \mathcal{\widehat{L}} + \sum_i \lambda_i \mathcal{\widehat{V}}_i$, we use the \textit{augmented Lagrangian method (ALM)}~\cite{powell1969method, hestenes1969multiplier}, which adapts $\lambda_i$ dynamically with violation magnitude:
}
\begin{equation}
\min_{\boldsymbol{\mu, \sigma}} \mathcal{L}_{total} = \mathcal{\widehat{L}} + \sum_i \left( \lambda_i \mathcal{\widehat{V}}_i + \frac{\rho}{2} \|\mathcal{\widehat{V}}_i\|^2 \right)
\end{equation}
where $\rho$ is a global penalty parameter.
Initially, $\lambda_i$ is set to a negligible value (e.g., $10^{-6}$).
After each iteration, $\lambda_i$ is updated according to the rule $\lambda \leftarrow \lambda_i + \rho \mathcal{\widehat{V}}_i$.
This mechanism ensures that hard constraints receive increasing attention from the optimizer until they are fully satisfied.

\input{algos/disco}
\textbf{Sampling}.
To sample a discrete schedule from the continuous representation, 
we round the mean ($\mu_i$) for each operator ($v_i$) to its nearest integer.
While ALM minimizes expected violations, the resulting rounded schedule may still contain legal conflicts.
To ensure the final output is valid, we apply a lightweight greedy algorithm to map illegal schedules to the nearest feasible ones.
These legalized schedules are then used to re-initialize the parameters ($\boldsymbol{\mu, \sigma}$) for further optimization (see Appendix \ref{app:legal} for detailed algorithm).

\textbf{Initialization}.
\new{
Unlike categorical models that start from a flat distribution, GauS can warm-start around heuristic solutions.
We initialize $\boldsymbol{\mu} = (\boldsymbol{s}^{ASAP} + \boldsymbol{s}^{ALAP}) / 2$, or a list-scheduling output, and $\boldsymbol{\sigma} = \kappa \cdot (\boldsymbol{s}^{ALAP} - \boldsymbol{s}^{ASAP})$ proportional to each node's scheduling freedom, where $\kappa$ controls its scale.
}

To utilize the parallelism of modern GPUs, we vectorize the entire workflow.
The complete algorithm for GauS is summarized in Algorithm \ref{alg:ours}.

%% file: figures/gaussian.tex
\begin{wrapfigure}{r}{0.5 \linewidth}
    \centering
    \includegraphics[trim={2cm 0 1.2cm 0.5cm}, clip, width=\linewidth]{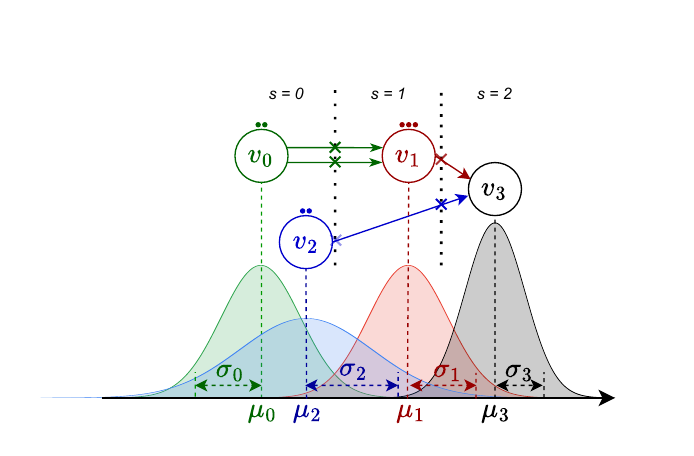}
    \caption{
    \textbf{Illustration of Gaussian reparameterization} ---
    \small{
    Each operator $v_i$ is parameterized as an independent Gaussian $X_i \sim \mathcal{N}(\mu_i, \sigma_i^2)$ over the scheduling timeline.
    }}
    \label{fig:gaussian}
    \vspace{-30pt}
\end{wrapfigure}

%% file: algos/disco.tex
\begin{wrapfigure}{r}{0.5\linewidth}
\begin{minipage}{\linewidth}
\begin{algorithm}[H]
   \caption{GauS algorithm}
   \label{alg:ours}
\begin{algorithmic}[1]
   \STATE {\bfseries Input:} $G = (V, E)$, max depth $D$, initial mean $\boldsymbol{\mu}_0$
   \STATE {\bfseries Parameters:} Lagrange ratios $\rho$, std factor $\kappa$.
  \STATE \textbf{Initialization:}
   \STATE \quad Compute $\boldsymbol{s}^{ASAP}$ and $\boldsymbol{s}^{ALAP}$
   \STATE \quad Set $\boldsymbol{\sigma} = \kappa \cdot (\boldsymbol{s}^{ALAP} - \boldsymbol{s}^{ASAP})$
   \STATE \quad Set $\boldsymbol{\mu} = \boldsymbol{\mu}_0$
   \REPEAT
   \STATE Compute primary objectives $\mathcal{\widehat{L}}(\boldsymbol{\mu}, \boldsymbol{\sigma})$
   \STATE Compute expected constraint violations $\mathcal{\widehat{V}}_i (\boldsymbol{\mu}, \boldsymbol{\sigma}), \forall i$ 
   \STATE $\mathcal{L}_{total} \leftarrow \mathcal{\widehat{L}} + \sum_i\left(\lambda_i \mathcal{\widehat{V}}_i + \frac{\rho}{2} \|\mathcal{\widehat{V}}_i\|^2 \right)$
   \STATE Update $\boldsymbol{\mu}, \boldsymbol{\sigma} \leftarrow \text{optimizer}(\nabla \mathcal{L}_{total})$
   \STATE Update $\lambda_i \leftarrow \lambda_i + \rho \mathcal{\widehat{V}}_i, \forall i$
   \STATE Discrete schedule $\boldsymbol{s} = \text{round}(\boldsymbol{\mu})$
   \STATE $\boldsymbol{\mu}, \boldsymbol{\sigma} \leftarrow \text{legalize}(\boldsymbol{s})$ \quad \textbf{if} $\boldsymbol{s}$ is illegal
   \UNTIL{convergence or time limit exceeded}
   \STATE {\bfseries Output:} Discrete schedule $\boldsymbol{s}$
\end{algorithmic}
\end{algorithm}
\end{minipage}
\end{wrapfigure}

%% file: 4-evaluation.tex
\section{Evaluation}
\subsection{Experimental Setup}

\input{figures/formA}
To ensure a fair comparison, a 15-minute time limit is applied to all methods.
Because GauS is deterministic given a fixed initialization, there is no variance between runs. 
Thus we report results from a single run per problem.
We implement GauS in PyTorch and use A100 GPUs with 80 GB memory to run the experiments.
More detailed settings are provided in Appendix \ref{app:exp}. 
\new{An ablation in Appendix~\ref{app:hp} confirms that GauS is robust to the choice of hyperparameters.}
\subsubsection{Problem Formulations}
Depending on the applications and context, scheduling problem may have different formulations~\cite{lam1988software, cong2006efficient, zhang2013sdc}.
In the evaluation part, we will focus on three distinct formulations.

\hypertarget{formA}{\textbf{Formulation A:}} \textit{Latency-constrained, resources and communication optimization for regular scheduling} .
This formulation is used in GS-Schedule~\cite{liu2024differentiable} which suits communication bottle-necked applications.
\begin{equation}
    \min_\mathbf{{s}} \mathcal{L}_{Res} + \alpha \mathcal{L}_{Com} \quad\text{s.t. \underline{Dep}, \underline{Lat}}
\end{equation}
Here $\mathcal{L}_{Com}$ represents an auxiliary communication overhead described in Appendix \ref{app:com}, and $\alpha$ is an application-specific constant that balances the tradeoff between resources and communication, 
$\boldsymbol{s}\in \mathbb{N}^n$ is vector form of $s_i, i\in\{0, \dots, n-1\}$.

\hypertarget{formB}{\textbf{Formulation B:}} \textit{Latency-constrained, memory footprint optimization for regular scheduling}.
In some realistic hardware workloads~\cite{amaru2015epfl}, the peak storage units is more important:
\begin{equation}
    \min_\mathbf{{s}} \mathcal{L}_{Mem} \quad\text{s.t. \underline{Dep}, \underline{Lat}}
\end{equation}

\textbf{Formulation C:} \textit{Latency, resources, and recurrence-constrained, memory footprint optimization for modulo scheduling}.
To further demonstrate the flexibility, 
we also use modulo scheduling~\cite{rau1981some, lam1988software, soi2025optimal}.
For a given resources limit $C_{res}$, and given II:
\begin{equation}
    \min_\mathbf{{s}} \mathcal{L}_{MMem}\quad
    \text{s.t. } \mathcal{L}_{MRes} \le C_{Res}, \text{\underline{Dep}, \underline{Lat}, \underline{Rec}}
\end{equation}

\subsubsection{Benchmarks}
\input{figures/stats}
Due to the lack of publicly available large-scale scheduling datasets, we adopt and extend the benchmarks established in recent literature~\cite{liu2024differentiable}.
For unpipelined formulations (\hyperlink{formA}{A} and \hyperlink{formB}{B}), we evaluate GauS using the same suite as~\cite{liu2024differentiable}, which includes realistic workloads from the EPFL Benchmark Suites~\cite{amaru2015epfl} and {synthetic random workloads (RW).}


For the pipelined formulation \hyperlink{formC}{C}, existing benchmarks~\cite{zhang2013sdc, oppermann2019exact} are too small for modern requirements.
Thus, we synthesized new benchmarks by augmenting the EPFL and RW suites with resource and recurrence constraints, with the number of added recurrences proportionally to graph size.
The overview of problem size is shown in Figure \ref{fig:stats}. 
More details can be found in Appendix \ref{app:data}.

\subsubsection{Baselines}

\textbf{Exact Solvers}.
To understand the tradeoff on the quality spectrum,
we use the state-of-the-art SDC+LP formulation~\cite{cong2006efficient, zhang2013sdc} with the commercial solvers CPLEX~\cite{cplex} and Gurobi~\cite{gurobi} 
as baselines.

\textbf{Heuristics}.
To further understand the tradeoff on the speed spectrum, 
we also compare our methods with two popular heuristics: 
List Scheduling~\cite{parker1986maha, kondratyev2011realistic} and Force-Directed Scheduling (FDS)~\cite{verhaegh1992force, paulin1987force}.

\textbf{Differentiable method}.
We compare to GS-Schedule only on formulation \hyperlink{formA}{A}, because this is the original formulation used in their work and is not straightforward to extend GS-Schedule to other formulation.

\input{figures/formB}
\input{figures/formC}
\subsection{Quality Comparison}

\textbf{Formulation} \hyperlink{formA}{\textbf{A}}: 
The quality comparison between GauS and traditional combinatorial solvers (CPLEX, Gurobi) as well as GS-Schedule is illustrated in Figure \ref{fig:formA}.
The benchmarks are ordered by operator count ($|V|$) from left to right.

For smaller benchmarks, traditional ILP solvers often find optimal or near-optimal solutions.
As the problem scale increases, GauS consistently produces high-quality schedules within the 15-minute time limit.
In contrast, traditional solvers marked with light crosses fail to produce a feasible solution or result in a schedule quality exceeding $4\times$ the performance threshold relative to GauS.

While GS-Schedule is able to produce solutions on most benchmarks, 
the solution quality is about $30\%$ to $3\times$ worse compared to GauS with a geometric mean of $71.8\%$.
This demonstrates the superior efficacy of GauS.
The memory efficiency reflects the other advantage of Gaussian reparameterization, parameter efficiency.
As shown in the EPFL benchmarks, the GS-Schedule baseline frequently exceeds available CUDA memory (OOM) on larger graphs (represented by filled cross markers), 
while GauS is able to produce solutions for large graphs. 
This parameter explosion is a direct consequence of their $O(ND)$ parameter space,
whereas GauS maintains a linear parameter space, 
enabling the optimization of massive-scale designs that are physically untreatable by previous differentiable methods.

\textbf{Formulation \hyperlink{formB}{\textbf{B}} and \hyperlink{formC}{C}}: 
To demonstrate the versatility of GauS, we evaluate its performance on two advanced scheduling formulations: \textit{latency-constrained, memory footprint optimization} and \textit{latency, resource, and recurrence-constrained modulo optimization}.

While exact solvers like CPLEX and Gurobi achieve competitive results on smaller graphs, their performance degrades as problem scale increases, frequently exceeding the 15-minute time limit or failing to produce any feasible solution (indicated by light crosses).
In contrast, GauS maintains a robust and stable performance profile, consistently matching or exceeding the quality of standard heuristics such as List Scheduling and Force-Directed Scheduling (FDS).
On most benchmarks, these heuristics often result in $20\%$ to $60\%$ higher memory footprint than our method.

The performance gap narrows in large graphs in formulation \hyperlink{formC}{C} on EPFL. 
As the number of augmented back-edges is proportional to the graph size,
feasible search space is significantly restricted, especially for large graphs.
decreasing the overall degrees of freedom,
making the margin for potential improvement becomes inherently smaller.






\subsection{Tradeoff Analysis}
\input{figures/tradeoff}

To evaluate optimization efficiency, we analyze the anytime results (solution quality vs. runtime) for GauS against baselines.
Figure \ref{fig:tradeoff} illustrates tradeoff for a subset of problems in Formulation A, with comprehensive results for all benchmarks provided in Appendix \ref{app:tradeoff}.
GauS consistently reaches high-quality solutions significantly faster than all baselines.
even for RW\_5 where the eventual solution slightly falls short.
When both methods eventually converge to similar objectives,
GauS achieves equivalent solution quality one to two orders of magnitude faster.
The results show that GauS achieves Pareto-optimal results.

Furthermore, GauS significantly enhances the utilization of GPU parallelism as detailed in Appendix \ref{app:gpu}.
While GS-Schedule drops below 40\% on averaged GPU utilization as complexity increases, GauS maintains near-100\%.

%% file: figures/formA.tex
\begin{figure*}[htbp]
    \centering
    \includegraphics[width=\textwidth]{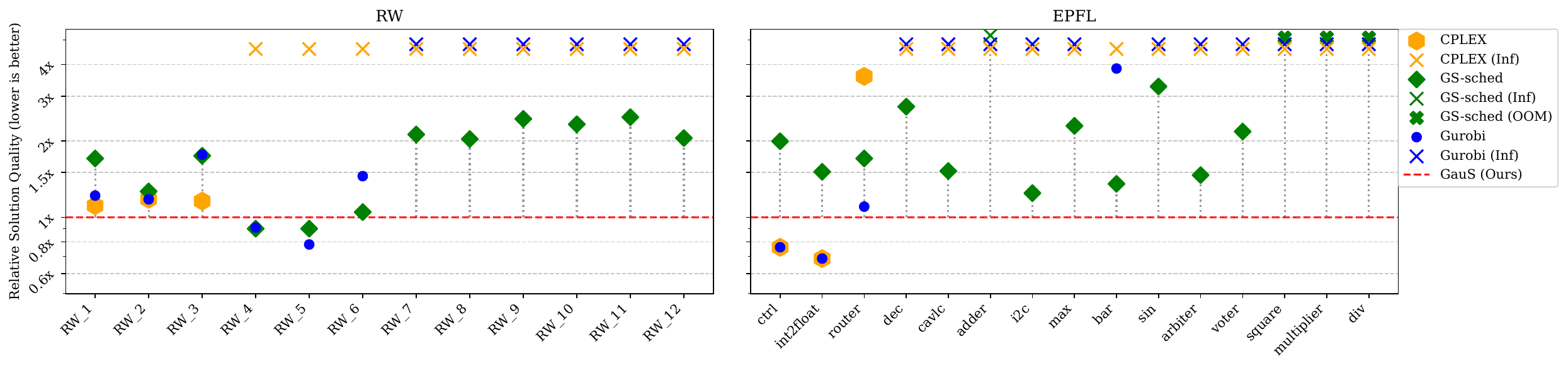}
    \caption{\hyperlink{formA}{\textbf{Formulation A: }} \textit{Latency constrained, resource and communication optimization} --- \small{
    Problem instance size ($|V|$) increases from left to right.
    A 15-minute time limit is applied to all methods.
    Relative solution quality is represented as the ratio between the methods' solution quality and GauS, where lower values correspond to higher solution quality.
    Light crosses (labeled as inf) denote no feasible solution or quality exceeding $4\times$ GauS performance;
    filled crosses indicate GS-Schedule exceeded available CUDA memory (OOM).
    On shared feasible instances, GauS achieves a $\textbf{71.8\%}$ geometric mean improvement over GS-Schedule.
    }}
\label{fig:formA}
\end{figure*}

%% file: figures/stats.tex
\begin{wrapfigure}{r}{0.5\linewidth}
    \centering
    \includegraphics[width=0.9\linewidth]{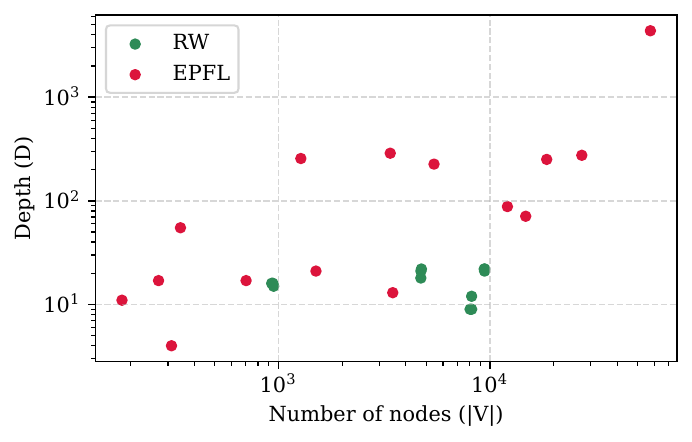}
    \caption{
    \textbf{Benchmark statistics}
    }
    \label{fig:stats}
    \vspace{-10pt}
\end{wrapfigure}

%% file: figures/formB.tex
\begin{figure*}[t]
    \centering
    \includegraphics[width=\textwidth]{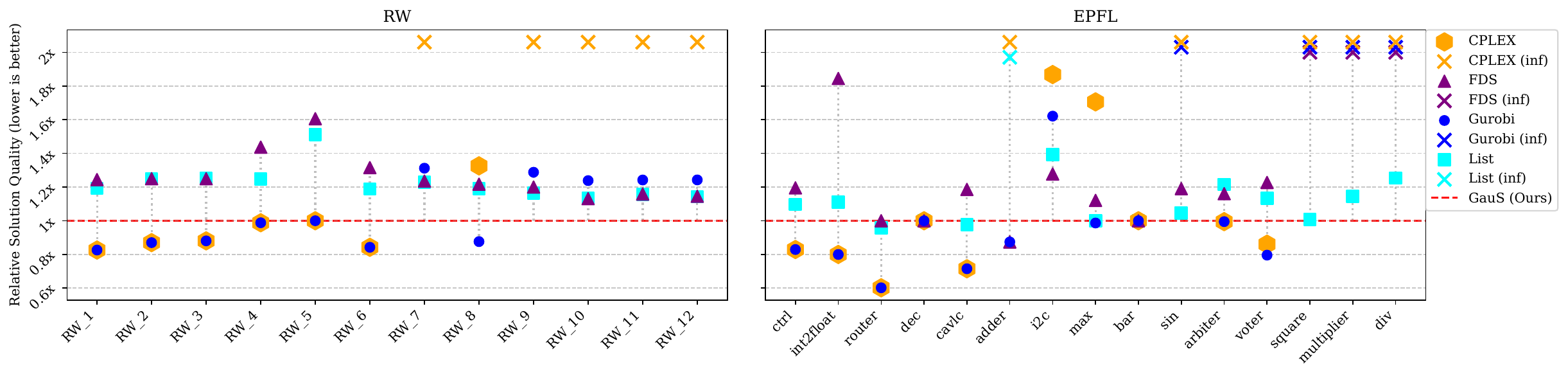}
    \caption{\hyperlink{formB}{\textbf{Formulation B: }} \textit{Latency constrained, memory footprint optimization} --- 
    \small{A 15-minute time limit is applied to all methods.
    Light crosses denote no feasible solution or quality exceeding $4\times$ GauS performance.
}}
    \label{fig:formB}
\end{figure*}

%% file: figures/formC.tex
\begin{figure*}[t]
    \centering
    \includegraphics[width=\textwidth]{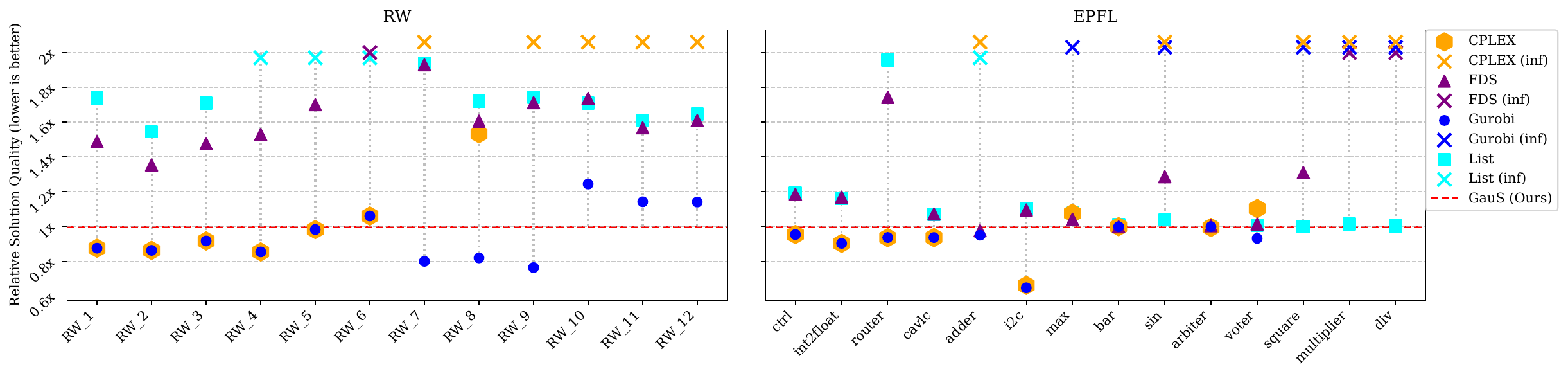}
    \caption{\hyperlink{formC}{\textbf{Formulation C: }} \textit{Latency, resources, and recurrence-constrained, memory footprint modulo optimization} --- 
    \small{A 15-minute time limit is applied to all methods.
    Light crosses denote no feasible solution or quality exceeding $4\times$ GauS performance.
    dec in EPFL has no feasible solution, thus omitted from the plot.
    }}
    \label{fig:formC}
\end{figure*}

%% file: figures/tradeoff.tex
\begin{figure*}[t!]
    \centering
    \includegraphics[width=\linewidth]{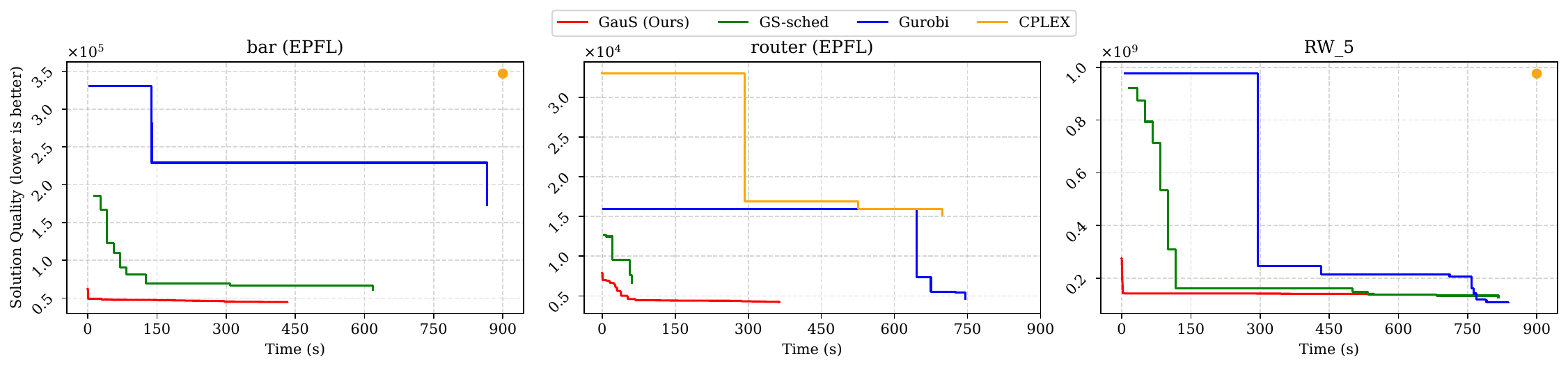}
    \caption{\textbf{Quality-Speed tradeoff representative benchmarks} -- 
    From left to right: RW\_5, router, and bar on formulation \hyperlink{formA}{A}. 
    Lines terminating before the 15-minute indicate no further improvement is made. }
    \label{fig:tradeoff}
\end{figure*}

%% file: 5-conclusion.tex
\section{Conclusion}
In this work, we presented GauS, a scalable differentiable framework for operator scheduling designed to overcome the bottlenecks of traditional combinatorial solvers and categorical differentiable models.
By leveraging a Gaussian reparameterization, GauS achieves $O(|V|)$ parameter scalability, enabling the optimization of massive-scale hardware designs with tens of thousands of nodes that were previously considered intractable.
As the first differentiable method to support pipelined scheduling, GauS offers a unique level of flexibility.
Our experimental results across realistic and synthetic workloads demonstrate that GauS delivers the state-of-the-art convergence speeds and high-quality solutions, making it as a robust tool for modern scheduling problem.

Despite its scalability, the current framework may occasionally converge to suboptimal solutions.
This is partly due to the modeling of operators as independent Gaussian variables, which neglects interactions between nodes, particularly those sharing the same edge.
While a Gaussian Process could capture these correlations, it introduces a prohibitive $O(|V|^2)$ space complexity.
Future work could explore more efficient ways to model node correlations and investigate advanced optimization strategies to further enhance convergence stability and solution quality.

%% file: a1-appendix.tex
\newpage
\appendix
\onecolumn

\section{Legalization Heuristic}
\label{app:legal}

While the Gaussian stochastic relaxation encourages valid schedules through dependency penalties, rounding learned means $\mu_i$ to discrete integers $s_i$ can occasionally result in causality violations.
Therefore, we use some heuristics to legalize any solution that violates the constraint. 
Because of the recurrence constraint in modulo scheduling.
we use two different legalization heuristic for regular and modulo scheduling problem.

\subsection{Regular Scheduling Legalization Heuristic}
To ensure strict feasibility, we employ a topological legalization pass.
This procedure processes the graph in topological order, shifting nodes forward in time only when necessary to satisfy dependency latencies while remaining within the pre-defined ASAP and ALAP bounds.
By performing this one-pass refinement, we try to legalize the discrete output without significantly altering the global optimization results.
The algorithm is summarized in Algorithm \ref{alg:legalize}

\begin{algorithm}[hb]
\caption{Regular Schedule Legalization}\label{alg:legalize}
\begin{algorithmic}
\STATE {\bfseries Input:} $G=(V, E)$, rounded schedule $s$.
\STATE Compute $\boldsymbol{s}^{ASAP}$ and $\boldsymbol{s}^{ALAP}$
\STATE $\boldsymbol{s}_{new} \leftarrow \text{clamp}(\boldsymbol{s}, \text{min}=\boldsymbol{s}^{ASAP}, \text{max}=\boldsymbol{s}^{ALAP})$
\FOR{{each} node $v$ {in} topological\_sort($G$)}
    \STATE $preds \leftarrow \{v_j \mid (v_j, v) \in E\}$
    \IF{$|preds| > 0$}
        \STATE $t_{req} \leftarrow \max(\boldsymbol{s}_{new}[Preds])+1$
        \STATE $s_{new}[v] \leftarrow \max(\boldsymbol{s}_{new}[v], t_{req})$
    \ENDIF
\ENDFOR
\STATE {\bfseries Output:} Feasible discrete schedule $s_{new}$.
\end{algorithmic}
\end{algorithm}

\subsection{Modulo Scheduling Legalization Heuristic}
Modulo scheduling introduces cyclic constraints due to back-edges.
To resolve these, we employ a simple fixed-point relaxation algorithm.
Unlike standard topological legalization, this iterative approach repeatedly passes through the graph to resolve back-edge requirements alongside forward dependencies.
The process continues until the schedule stabilizes or a maximum iteration threshold (equal to the number of nodes) is reached, signaling a potential recurrence violation or depth overflow. 
This tries to find a feasible schedule for pipelined execution while not significantly changing the original input. 
The algorithm is summarized in Algorithm \ref{alg:modulo_legalize}.

\begin{algorithm}[ht]\caption{Modulo Schedule Legalization}\label{alg:modulo_legalize}
\begin{algorithmic}
\STATE {\bfseries Input:} $G=(V, E)$, back-edge set $E_B$, max depth $D$, initiation interval \text{II}, rounded schedule $\boldsymbol{s}$.
\STATE $\boldsymbol{s}_{new} \leftarrow \boldsymbol{s}$
\STATE $nodes \leftarrow \text{topological\_sort}(G)$
\FOR{$i=0$ {\bfseries to} $|V|-1$}
    \STATE $changed \leftarrow \text{False}$
    \FOR{{\bfseries each} $v$ {\bfseries in} $nodes$}
        \STATE $t_{min} \leftarrow \max(\{\boldsymbol{s}_{new}[u] \mid u \in preds(v)\} \cup \{0\}) + 1$
        \STATE $t_{back} \leftarrow \max(\{\boldsymbol{s}_{new}[v_j] - k\cdot \text{II} + 1 \mid (v, v_j, k) \in E_B\} \cup \{0\})$
        \STATE $t_{req} \leftarrow \max(t_{min}, t_{back})$
        \IF{$t_{req} > \boldsymbol{s}_{new}[v]$}
            \STATE $\boldsymbol{s}_{new}[v] \leftarrow t_{req}, \quad changed \leftarrow \text{True}$
        \ENDIF
    \ENDFOR
    \IF{not changed}  
        \STATE \textbf{break} 
    \ENDIF
\ENDFOR
\STATE {\bfseries Output:} $s_{new}$.
\end{algorithmic}
\end{algorithm}

\subsection{\new{Empirical Behavior of Regular Legalization}}
\new{To verify that legalization acts as a feasibility safety net rather than a substitute for the optimizer, we suppressed the legalization algorithm and ran a sweep of all EPFL benchmarks under the formulation A.
Every scheme produces a feasible schedule on all benchmarks.
Across the all feasible runs, the legalization is invoked on 75--95\% of runs and roughly twice per run on average, but the legalized schedule improves on the running best in only a small fraction of events (below 15\%); in the remainder, the optimizer is reset to the prior best.
The typical perturbation when legalization fires is modest (median $\sim$10--20\% of nodes shifted), with a heavy tail of occasional events re-seating the majority of nodes, confirming that legalization is overwhelmingly a safety net rather than a primary source of optimization signal.}

\section{Derivations}
\label{app:drv}
We finish the derivation of expected objectives and violations of constraints in this section. 
After defining communication overhead, we will compute its expectation and expected resource usage under regular scheduling before deriving the metrics essential to modulo scheduling

\subsection{Communication overhead}
\label{app:com}
In communication-bottlenecked workloads~\cite{zhang2022gatspi}, we minimize the total edge length across all dependencies:
\begin{equation}
    \mathcal{L}_{Com} = \sum_{(v_i, v_j)\in E} s_j - s_i
\end{equation}

Then the expected total edge length $\widehat{\mathcal{L}}_{Com}$ is the sum of all expected violations from all edges:
\begin{equation}
    \mathcal{\widehat{L}}_{Com} = \mathbb{E}[\mathcal{L}_{Com}] 
     = \sum_{(v_i, v_j) \in E} \sum_{d_i=0}^{D-1}\sum_{d_j=d_i}^{D-1} P_i^{d_i}P_j^{d_j} \cdot(d_j - d_i) \\
\end{equation}

\subsection{Expected Resource Usage}
To represent the expected resource usage at step $d$, we reuse the probability that an operator $v_i$ with resource demand $w_i$ is active at that step defined in Equation \ref{eq:def}.
The expected resource usage is the sum of individual usage:
\begin{equation}
\widehat{Res}(d) = \mathbb{E}[Res(d)] = \sum_{v_i \in V} w_i \cdot P_i^d
\end{equation}
To optimize for peak resource utilization, we apply the \texttt{LogSumExp} operator across all steps $d \in \{0, \dots, D-1\}$.
This is similar to the computation of global memory footprint in \ref{sec:drv}:
\begin{equation}
\widehat{\mathcal{L}}_{Res} \approx \tau\cdot  \log \left( \sum_{d} \exp \left(\frac{ {\widehat{Res}(d)}}{\tau}\right) \right)
\end{equation}

We define the violations as the number of exceeded resources used summed across all steps.
Then the expected violations can be computed as:
\begin{equation}
    \widehat{\mathcal{V}}_{Res} = \sum_d \text{ReLU}\left( \widehat{{Res}}(d) -  R \right)
    \label{eq:relu}
\end{equation}
where $R$ is a given resources limit, and $\text{ReLU}(\cdot)$ ensures only exceeding resources are counted toward violations.

\subsection{Modulo Metrics}
In a pipeline, an operator scheduled at step $d_i$ consumes resources at steps $d_i \mod{II}$.
The probability that operator $v_i$ occupies a resource at time $t \in \{0, \dots, II-1\}$ within the modulo reservation table is the sum of its probabilities across all linear steps that map to $t$:
\begin{equation}
P\left\{\lfloor X_i +0.5 \rfloor = t \right\} = \sum_{k=0}^{\lceil D/\text{II} \rceil} P_i^{t + k \cdot II}
\end{equation}
where $P_i^{t+k\cdot \text{II}}$ represents the probability of scheduling operator $v_i$ to step $d=t+k\cdot \text{II}$ is defined in Equation \ref{eq:round}.

\textbf{Modulo Resource Usage}. 
Similar to the derivation in \ref{sec:drv},
at step $0\le t<\text{II}$, the expected modulo resource usage $\widehat{MRes}(t)$:
\begin{equation}
\widehat{MRes}(t) = \sum_{v_i \in V} w_i \cdot P\left\{\lfloor X_i +0.5 \rfloor = t \right\}
\end{equation}

\textbf{Modulo Memory Footprint}.
Memory footprint in a pipeline is similarly wrapped.
A memory unit carrying the output of $v_i$ is active at step $t$ if it is active at any time step $d$ such that $d = t+k\cdot\text{II}, k>0$.
Using the linear probability $P(X_i \le s < \max_{v_j \in succ(v_i)} X_{j})$ defined in Equation \ref{eq:active}, we can compute the modulo expectation:
\begin{equation}
\widehat{MReg}(t) = \sum_{v_i \in V} b_i \cdot \sum_{k=0}^{\lceil D/\text{II}\rceil} P\left\{d_i \le t + k\cdot\text{II} < \max_{v_j \in{succ(i)}} d_j\right\}
\end{equation}

The expected maximum modulo resource usage and peak memory footprint can be computed using \texttt{LogSumExp} similar to Equation \ref{eq:max}.
The expected violations can be computed using ReLU similar to Equation \ref{eq:relu}.

\textbf{Recurrence Violations}.
For modulo scheduling with loop-carried dependencies, a back-edge from producer $v_j$ to consumer $v_i$ with an iteration distance $k$ and latency Lat$(v_i)$ imposes the constraint $s_i + k \cdot \text{II} \ge s_j + \text{Lat}(v_i)$.
Let the set of these dependencies be defined as $(v_j, v_i, k) \in E_B$.
Similar to the computation of expected dependency violations in Equation \ref{eq:dep}, 
the expected recurrence violation $\widehat{\mathcal{V}}_{rec}$ is formulated by summing the probabilities of all invalid steps assignments $(d_i, d_j)$ that fail to satisfy the recurrence constraint:
\begin{equation}
    \mathcal{\widehat{V}}_{Rec} = \mathbb{E}[\mathcal{V}_{Rec}]  
    = \sum_{(v_i, v_j, k) \in E_B} \sum_{d_i=1}^{D-1} \sum_{d_j=d_i + k\cdot \text{II} + 1}^{D-1} P_i^{d_i} \cdot P_j^{d_j} 
\end{equation}

\section{Detailed Evaluation Setup}
\subsection{Experiment Setup}
\label{app:exp}
\textbf{Environment Settings}.
All the experiments are performed on a Linux server equipped two NVIDIA A100 GPUs with 80 GB memory and two AMD EPYC 9124 CPUs (2$\times$16 cores) running at 3.7 GHz, 1.5 TB of RAM.
For softwares, we use PyTorch 2.5.1 with CUDA 12.1, Gurobi 13.0, and CPLEX 22.1.1.

\textbf{Hyper-parameter Settings}.
We use a uniform set of hyperparameters across all problem instances and formulations.
We use the Adam optimizer with a learning rate of $10^{-2}$.
The penalty parameter $\rho$, used to enforce constraints, is set to $10^{-4}$;
the \texttt{LogSumExp} temperature parameter $\tau$ is set to $10^{-2}$;
the initialization freedom parameter $\kappa$ is set to $\frac{1}{6}$.

\subsection{Dataset Details}
\label{app:data}
For the random workload (RW) evaluations, we utilize a subset of the datasets released by GS-Schedule~\cite{liu2024differentiable},
though our selected subset is substantially larger than the one featured in the original study to better evaluate scalability.
\input{tables/stats}

\section{\new{Hyperparameter Sensitivity}}
\label{app:hp}

\new{GauS exposes three core hyperparameters that we hold fixed across every benchmark in the main paper: the initialization freedom $\kappa$, which scales each node's standard deviation $\sigma_i$ to a fraction $\kappa$ of its $[\textrm{ASAP},\,\textrm{ALAP}]$ slack; the \texttt{LogSumExp} temperature $\tau$ that smooths the resource peak; and the Adam learning rate. We sweep each in turn while holding the other two at their defaults ($\kappa{=}1/6$, $\tau{=}10^{-2}$, $\textrm{lr}{=}10^{-2}$) on five EPFL benchmarks (\texttt{dec}, \texttt{cavlc}, \texttt{bar}, \texttt{arbiter}, \texttt{multiplier}) covering three orders of magnitude in graph size, with three seeds per cell, yielding $254/255$ feasible runs within the $900$\,s time limit.}

\input{figures/hp_sensitivity}

\new{The default $\kappa = 1/6$ follows directly from the $3\sigma$ rule for the Gaussian. With $\sigma_i = \kappa\,(s_i^{\,\textrm{ALAP}}-s_i^{\,\textrm{ASAP}})$, taking $\kappa = 1/6$ places $\pm 3\sigma$ exactly at the boundary of the feasible $[\textrm{ASAP},\,\textrm{ALAP}]$ window for each node, so the initial distribution covers ${\approx}99.7\%$ of the feasible steps without leaking probability mass outside. Smaller $\kappa$ under-explores; larger $\kappa$ wastes mass on infeasible steps, which Figure~\ref{fig:hp_sensitivity} (left) confirms causes a measurable cost penalty --- particularly on shallow graphs (e.g.\ \texttt{dec}, depth ${\approx}4$) where the slack window is narrow and $\kappa{=}1$ pushes ${\approx}30\%$ of the initial mass outside it.}

\new{The other two hyperparameters are far less sensitive. The temperature $\tau$ keeps the consensus-normalized cost ratio under $1.04\times$ over $\tau \in [10^{-3}, 10^{0}]$ and only deteriorates at $\tau{=}10$, where the soft-max becomes nearly flat and provides little gradient signal about the actual peak resource step. The learning rate exhibits a shallow basin from $10^{-3}$ to $10^{-2}$ within $1.04\times$ of the consensus best, and degrades only at the extremes ($1.20\times$ at $\textrm{lr}{=}10^{-1}$). The chosen $\textrm{lr}{=}10^{-2}$ trades a small loss in average objective for faster convergence on the larger benchmarks, where lower learning rates do not finish within the $100\,000$-iteration cap. We note that this is a one-at-a-time study and HP interactions are not isolated; nonetheless the absence of feasibility cliffs across all $255$ cells supports the claim that GauS is robust to the choice of HPs around the defaults.}

\section{\new{Distribution-Family Ablation}}
\label{app:dist}

\new{The Gaussian choice in GauS is motivated by unimodality, smooth temporal-proximity gradients, and a differentiable CDF --- none of which are uniquely Gaussian. The forward pass only ever invokes the CDF, and no derivation in this paper requires Gaussianity. 
To test whether the framework's success is specific to the Gaussian or general to \textit{smooth symmetric unimodal CDFs}, we replace $\Phi(\cdot)$ in Equation~\ref{eq:round} with the logistic, Laplace, or Cauchy CDF, calibrating the scale parameter so that the same $\kappa = 1/6$ yields a comparable initial spread across families: variance-matching for logistic ($s = \sigma\sqrt{3}/\pi$) and Laplace ($b = \sigma/\sqrt{2}$); FWHM(Full width at half maximum)-matching for Cauchy ($\gamma = \sigma\sqrt{2\ln 2}$), since the Cauchy distribution has no finite variance. All other hyperparameters are held at the Form A defaults; we do not retune per family.}

\input{tables/dist}

\new{Across the $15$ EPFL benchmarks where all four families are feasible, three of four families fall within ${\sim}8\%$ on geometric-mean cost.
We read this as evidence that GauS is not specific to the Gaussian: any smooth symmetric unimodal CDF yields comparable optimization quality, and the choice of family is a tuning knob rather than a structural assumption.}

\section{\new{Temporal-Proximity Ablation}}
\label{app:temporal}

\new{Section~\ref{sec:reparam} argues that a key advantage of the Gaussian parameterization over a categorical one is the \emph{temporal-proximity} inductive bias: as $\mu_i$ shifts, the per-step probability mass $P_i^d$ moves smoothly between \emph{adjacent} time steps.
We isolate this mechanism with a controlled ablation.
Once $P_i^d$ is computed according to \ref{eq:round}, we apply a per-node fixed permutation of the $d$-axis (each node draws its own permutation of $[0, D)$ once from the run seed and holds it constant). 
Each column of $P$ still sums to $1$, gradients still flow through $(\mu, \sigma)$, and the parameter count is unchanged --- but the row index no longer corresponds to \textit{actual time step}, so $\sum_d P_i^d$ no longer equals $\Pr(X_i \le d)$ and per-step resource sums no longer aggregate operators that occupy the same step.
A worse final cost under shuffling is direct evidence that the proximity mechanism contributed real optimization signal.}

\new{We run paired baseline and shuffled arms on all EPFL graphs under Formulation~\hyperlink{formA}{A}. 
Of the remaining $19$ graphs, shuffling makes the schedule \emph{worse} on all $15$. 
The averaged cost degradation is $\textbf{35\%}$.
The largest single degradations are \texttt{adder} ($10.87\times$) and \texttt{priority} ($5.98\times$).}

\new{The optimization-loss landscape under shuffling may look \textit{easier} --- per-row resource sums flatten when each node's mass is independently permuted --- but that lower optimization loss does not translate into a better discrete schedule.
This matches the proximity argument in Section~\ref{sec:reparam}: the smooth probability-mass-on-adjacent-steps structure is what makes the gradient point toward a real schedule improvement; removing it leaves the optimizer chasing a meaningless surrogate.}

\section{GPU Efficiency Analysis}
\label{app:gpu}
To evaluate the practical GPU efficiency of GauS, we performed a comparative GPU profiling analysis against the baseline, GS-Schedule, using the EPFL benchmarks.
Figure \ref{fig:profile} illustrates the peak GPU memory usage and compute utilization across various workloads.

\textbf{Memory Scalability}: GauS exhibits a remarkably consistent and relatively low memory footprint across all benchmarks, roughly linear with the size of the graph ($|V|$).
In contrast, GS-Schedule shows a more drastic growth in memory consumption as the size ($|V|$) and the depth ($D$) of the graph increases, resulting out-of-memory (OOM) issues on the largest designs like square, multiplier, and div.

\textbf{Compute Utilization}: Our Gaussian parameterization consistently maintains near-100\% averaged GPU utilization.
This demonstrates that the $O(N)$ parameter space is not only memory-efficient but also highly amenable to parallel execution.
Conversely, GS-Schedule struggles to saturate the GPU, with utilization dropping below 40\% as the graph complexity increases, likely due theirs poor GPU implementation.

\input{figures/profile}

\section{Full Tradeoff Results}
\label{app:tradeoff}
\subsection{Formulation A}
The full tradeoff analysis on formulation \hyperlink{formA}{A} is shown in Figure \ref{fig:formA_rw} and \ref{fig:formA_epfl}, 
GauS is on the Pareto frontier for most of the problem workloads, especially larger ones.

\input{figures/tradeoff_rw_formA}
\input{figures/tradeoff_epfl_formA}

\subsection{Formulation B and C}
The comprehensive tradeoff results on formulation \hyperlink{formB}{B} and \hyperlink{formC}{C} are shown in Figure \ref{fig:formB_rw}, \ref{fig:formB_epfl}, \ref{fig:formC_rw} and \ref{fig:formC_epfl}.
The results demonstrate that while exact solvers like CPLEX and Gurobi achieve high-quality solutions for smaller problem instances,
GauS is better in scalability, delivering good solutions on large-scale benchmarks.
For medium-sized cases, our method offers a highly competitive tradeoff, reaching high-quality solutions orders of magnitude faster than exact solvers.
Furthermore, GauS consistently outperforms fast heuristics, such as List Scheduling and FDS, which often usually leads to suboptimal results.

\input{figures/tradeoff_rw_formB}
\input{figures/tradeoff_epfl_formB}

\input{figures/tradeoff_rw_formC}
\input{figures/tradeoff_epfl_formC}

%% file: tables/stats.tex
\begin{table}[htbp]
\centering
\caption{Summary of Graph Statistics (Sorted by Nodes)}
\label{tab:graph_stats}
\begin{tabular}{lrr|lrr}
\toprule
\multicolumn{3}{c}{\textbf{EPFL}} & \multicolumn{3}{c}{\textbf{Random Workloads (RW)}} \\
\cmidrule(lr){1-3} \cmidrule(lr){4-6}
Name & Number of Nodes ($|V|$) & Depth ($D$) & Name & Number of Nodes ($|V|$) & Depth ($D$) \\
\midrule
ctrl & 182 & 11 & RW\_1 & 929 & 16 \\
int2float & 271 & 17 & RW\_2 & 941 & 16 \\
dec & 312 & 4 & RW\_3 & 949 & 15 \\
router & 344 & 55 & RW\_4 & 4713 & 18 \\
cavlc & 703 & 17 & RW\_5 & 4716 & 21 \\
adder & 1276 & 256 & RW\_6 & 4741 & 22 \\
i2c & 1504 & 21 & RW\_7 & 8058 & 9 \\
max & 3377 & 288 & RW\_8 & 8192 & 12 \\
bar & 3471 & 13 & RW\_9 & 8193 & 9 \\
sin & 5440 & 226 & RW\_10 & 9396 & 22 \\
arbiter & 12095 & 88 & RW\_11 & 9432 & 21 \\
voter & 14759 & 71 & RW\_12 & 9447 & 22 \\
square & 18550 & 251 &  &  &  \\
multiplier & 27190 & 275 &  &  &  \\
div & 57375 & 4373 &  &  &  \\
\bottomrule
\end{tabular}
\end{table}

%% file: figures/hp_sensitivity.tex
\begin{figure}[ht]
    \centering
    \includegraphics[width=\textwidth]{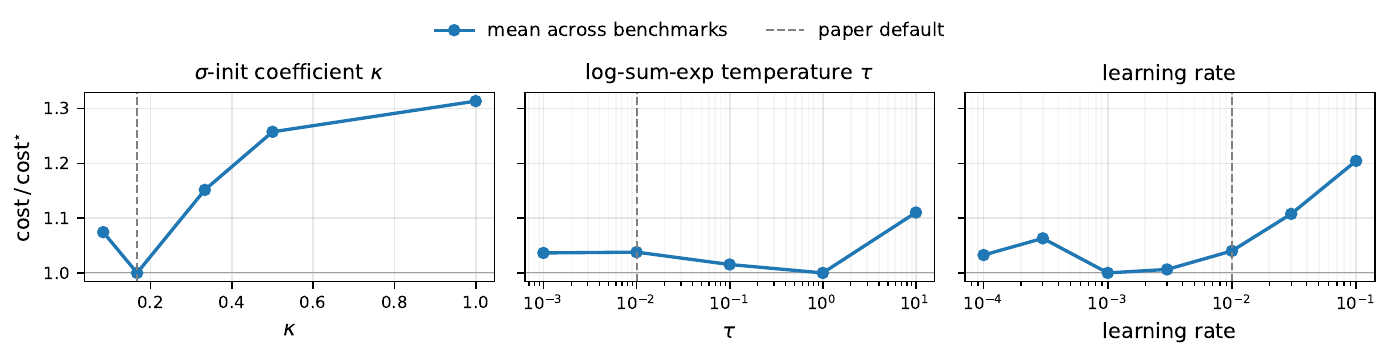}
    \caption{\new{Sensitivity of GauS to its three core hyperparameters under Formulation~A. Each panel sweeps one HP while the other two are held at the main-paper defaults ($\kappa{=}1/6$, $\tau{=}10^{-2}$, $\textrm{lr}{=}10^{-2}$). Per swept value we first compute, for each of five EPFL benchmarks (\texttt{dec}, \texttt{cavlc}, \texttt{bar}, \texttt{arbiter}, \texttt{multiplier}; 3 seeds each), the ratio of its mean cost to its own best across the swept range, and average those ratios across benchmarks. The curves are then re-normalized by their minimum so that $\mathrm{cost}/\mathrm{cost}^\star = 1$ at the consensus-best value. The dashed line marks the value used in the main paper: it coincides with the consensus best for $\kappa$, and lies within $1.04\times$ of it for $\tau$ and lr.}}
    \label{fig:hp_sensitivity}
\end{figure}

%% file: tables/dist.tex
\begin{table}[ht]
  \centering
  \new{
  \caption{Distribution-family ablation on EPFL (Form A, \texttt{icml} objective). Cells are mean cost relative to Gaussian ($\mathrm{cost}_{\mathrm{family}} / \mathrm{cost}_{\mathrm{Gaussian}}$); values below 1 indicate the family beats Gaussian. \textbf{Bold} marks the per-row winner. All four families share identical hyperparameters (no per-family retuning). Five benchmarks were run with 3 seeds and 14 with a single seed; \texttt{hyp} is excluded (load-time OOM for all four families).}
  \label{tab:dist}
  \begin{tabular}{lrrrrrl}
    \toprule
    Benchmark & seeds & Gaussian & Logistic & Laplace & Cauchy & Winner \\
    \midrule
    \texttt{ctrl} & 1 & \textbf{1.000} & 1.155 & 1.054 & 1.089 & Gaussian \\
    \texttt{int2float} & 1 & 1.000 & 1.007 & \textbf{0.994} & 0.998 & Laplace \\
    \texttt{router} & 1 & 1.000 & 1.006 & 0.939 & \textbf{0.895} & Cauchy \\
    \texttt{dec} & 3 & \textbf{1.000} & 1.400 & 1.302 & 1.661 & Gaussian \\
    \texttt{cavlc} & 3 & 1.000 & 1.004 & 1.019 & \textbf{0.952} & Cauchy \\
    \texttt{adder} & 1 & 1.000 & 1.112 & \textbf{0.534} & 1.334 & Laplace \\
    \texttt{i2c} & 1 & 1.000 & \textbf{0.845} & 0.853 & 0.975 & Logistic \\
    \texttt{max} & 1 & \textbf{1.000} & 1.115 & 1.119 & 1.044 & Gaussian \\
    \texttt{bar} & 3 & 1.000 & 0.946 & 0.921 & \textbf{0.786} & Cauchy \\
    \texttt{sin} & 1 & \textbf{1.000} & 1.383 & 1.389 & 1.461 & Gaussian \\
    \texttt{arbiter} & 3 & 1.000 & 1.142 & 1.157 & \textbf{0.984} & Cauchy \\
    \texttt{voter} & 1 & 1.000 & 0.974 & 0.950 & \textbf{0.944} & Cauchy \\
    \texttt{square} & 1 & \textbf{1.000} & 1.078 & 1.024 & 1.063 & Gaussian \\
    \texttt{multiplier} & 3 & \textbf{1.000} & 1.103 & 1.077 & 1.109 & Gaussian \\
    \texttt{div} & 1 & 1.000 & 0.999 & 1.000 & \textbf{0.997} & Cauchy \\
    \midrule
    Wins (out of 15) & --- & 6 & 1 & 2 & 6 & --- \\
    Geomean cost / Gaussian & --- & 1.000 & 1.075 & 1.002 & 1.066 & --- \\
    Mean rank (1 = best) & --- & 2.23 & 3.00 & 2.43 & 2.33 & --- \\
    \bottomrule
  \end{tabular}
  }
\end{table}

%% file: figures/profile.tex
\begin{figure}[ht]
    \centering
    \includegraphics[width=0.95\linewidth]{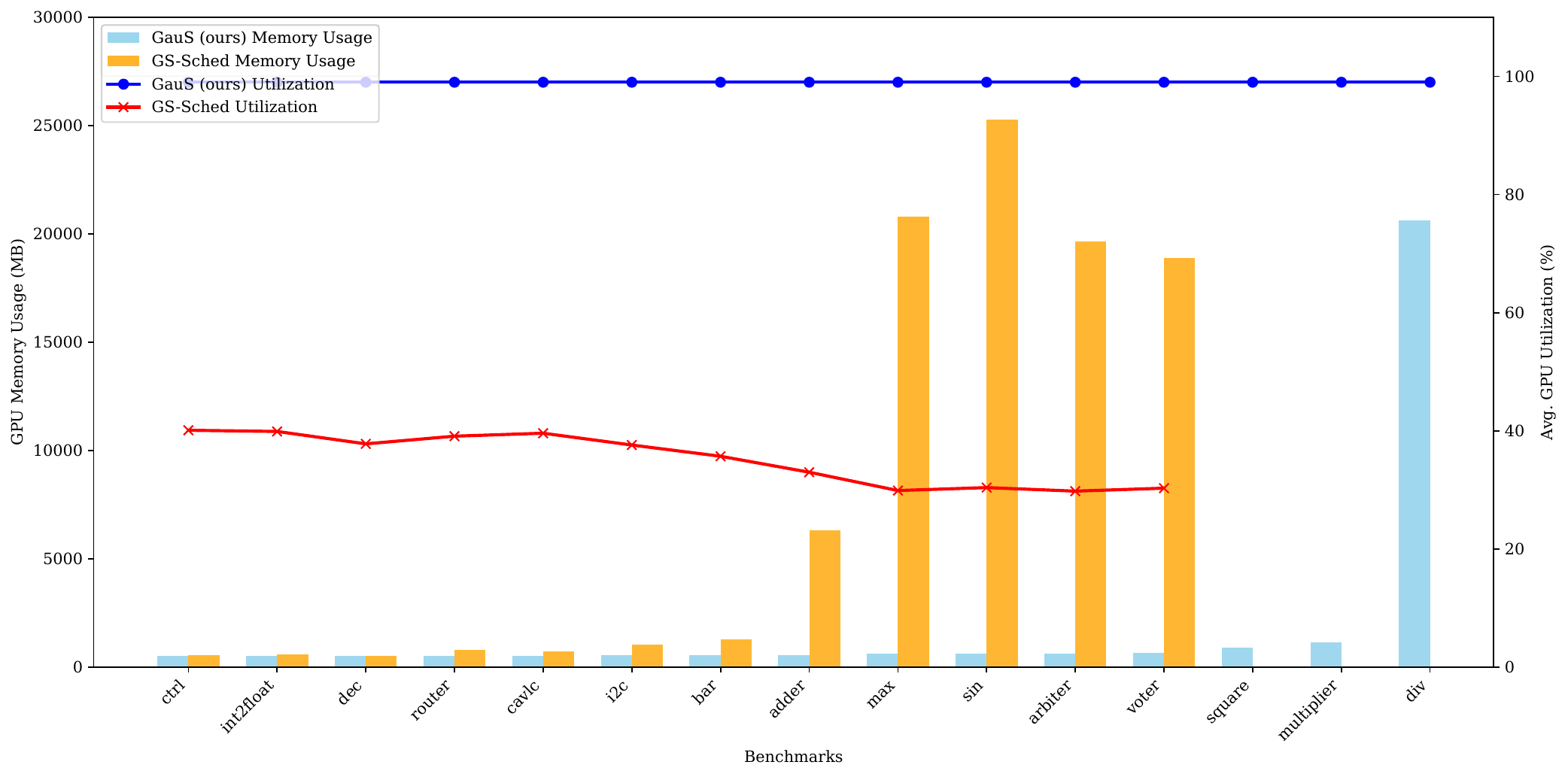}
    \caption{
    \textbf{GPU Profiling Results on EPFL} ---
    \small{The bar charts show the GPU memory usage, corresponding to the left axis. 
    The curve show the Avg. GPU utilization, corresponding to the right axis. 
    Avg. GPU utilization is the averaged GPU utilization (\%) during the entire execution. 
    Because GS-Schedule is out-of-memory (OOM) on square, multiplier, and div, 
    no data is provided for GS-Schedule on these benchmarks}
    }
    \label{fig:profile}
\end{figure}

%% file: figures/tradeoff_rw_formA.tex
\begin{figure*}[!ht]
    \centering
    \includegraphics[width=\textwidth]{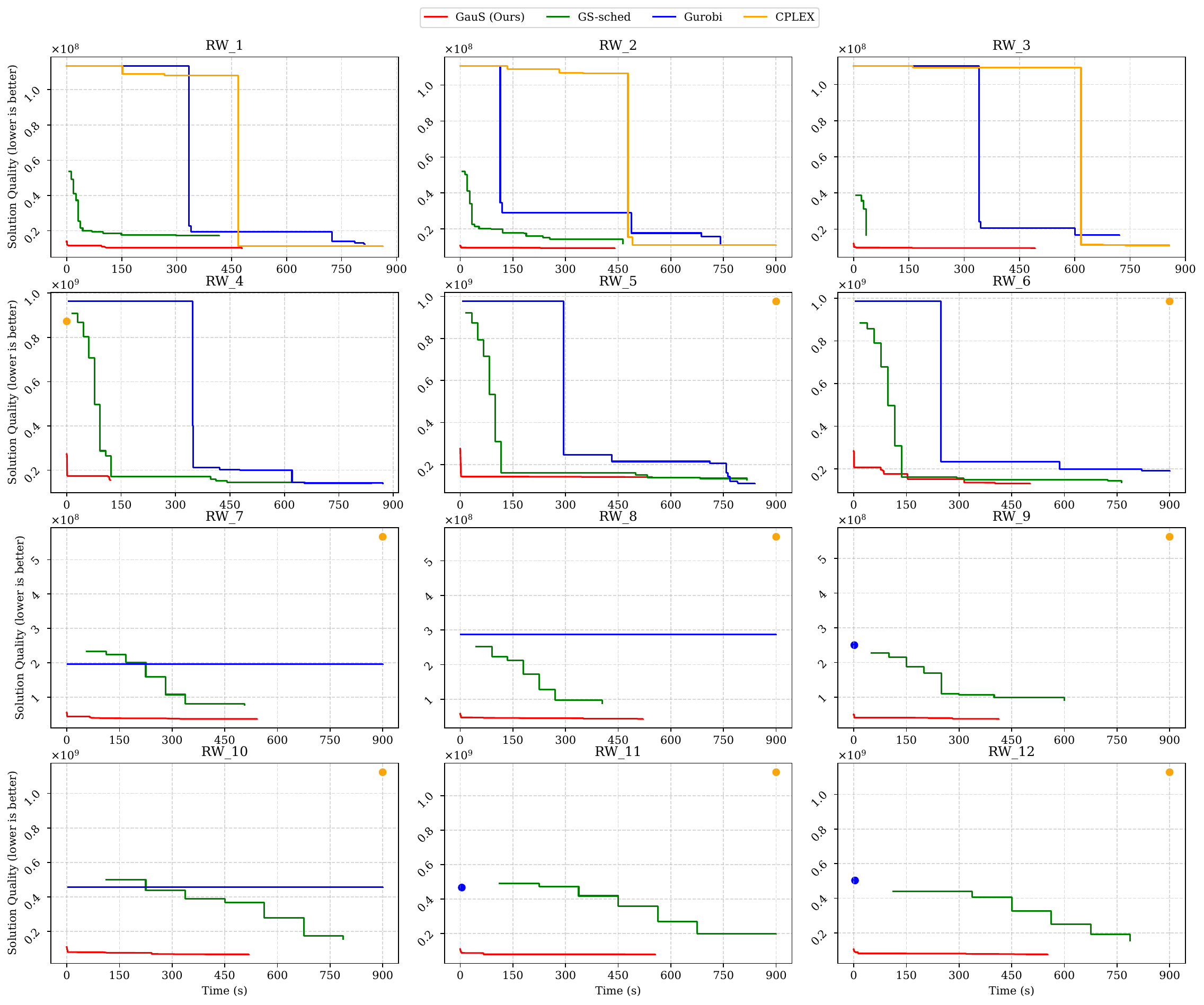}
    \caption{Comprehensive tradeoff comparison of RW on \hyperlink{formA}{\textbf{Formulation B: }} \textit{Latency constrained, resource and communication optimization} --- 
    \small{A 15-minute time limit is applied to all methods.
    Heuristics such as FDS and List Scheduling are represented as horizontal lines to reflect their fast convergence to final solution quality.
    }}
    \label{fig:formA_rw}
\end{figure*}

%% file: figures/tradeoff_epfl_formA.tex
\begin{figure*}[htbp]
    \centering
    \includegraphics[width=\textwidth]{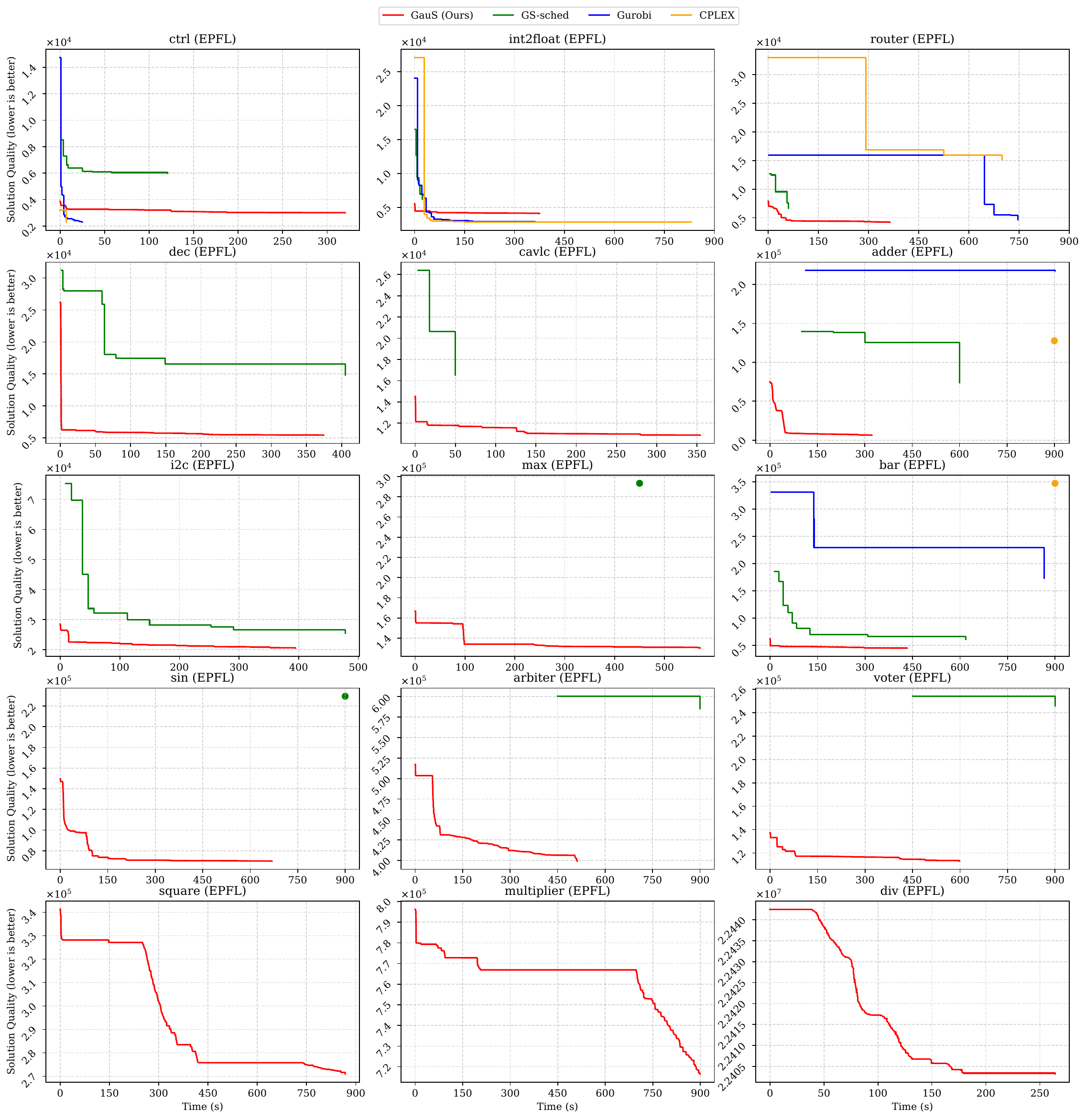}
    \caption{Comprehensive tradeoff comparison of EPFL on \hyperlink{formA}{\textbf{Formulation A: }} \textit{Latency constrained, resource and communication optimization} --- 
    \small{A 15-minute time limit is applied to all methods. }}
    \label{fig:formA_epfl}
\end{figure*}

%% file: figures/tradeoff_rw_formB.tex
\begin{figure*}[!ht]
    \centering
    \includegraphics[width=\textwidth]{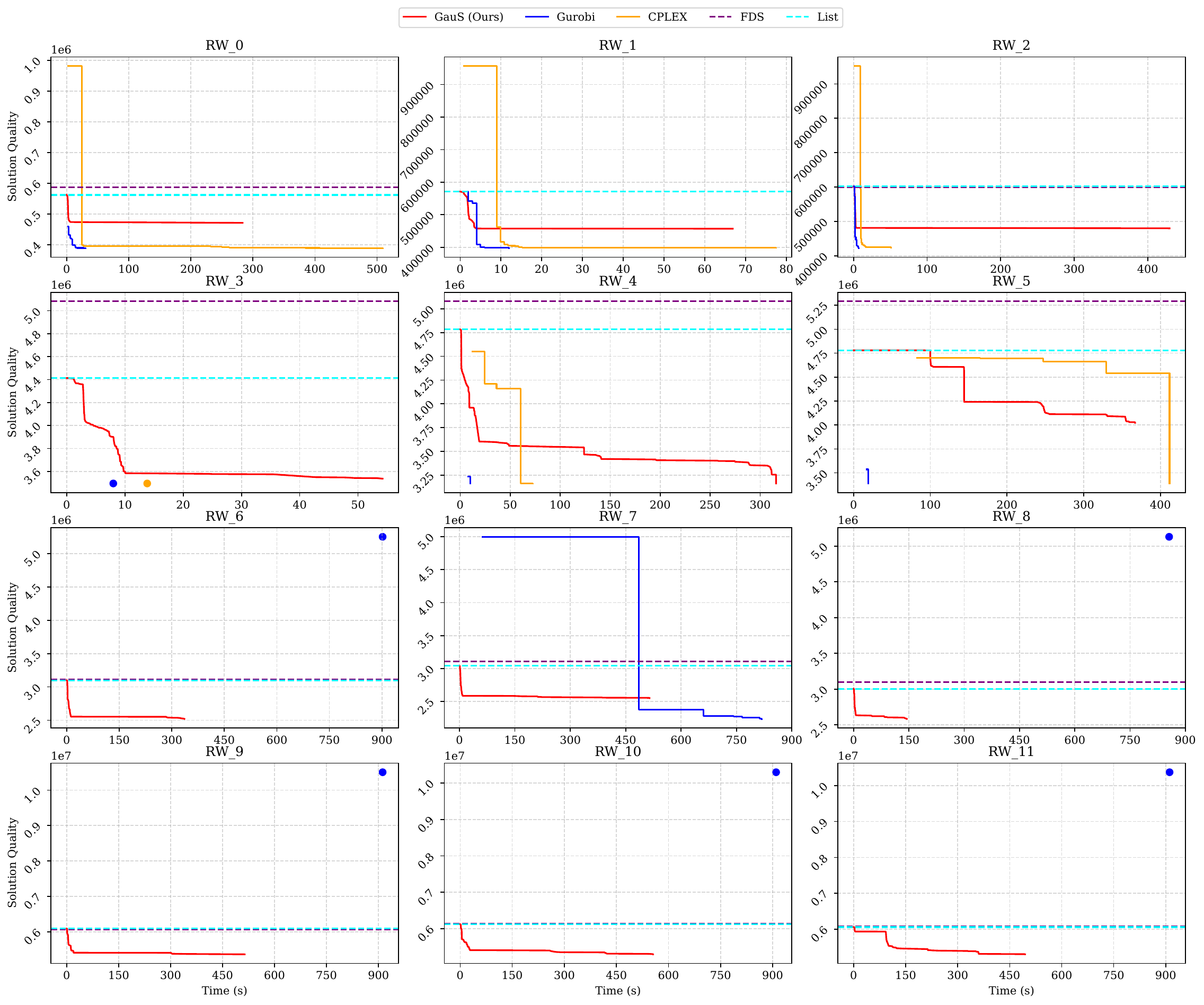}
    \caption{Comprehensive tradeoff comparison of RW on \hyperlink{formB}{\textbf{Formulation B: }} \textit{Latency constrained, memory footprint optimization} --- 
    \small{A 15-minute time limit is applied to all methods.
    Heuristics such as FDS and List Scheduling are represented as horizontal lines to reflect their fast convergence to final solution quality.
    }}
    \label{fig:formB_rw}
\end{figure*}

%% file: figures/tradeoff_epfl_formB.tex
\begin{figure*}[htbp]
    \centering
    \includegraphics[width=\textwidth]{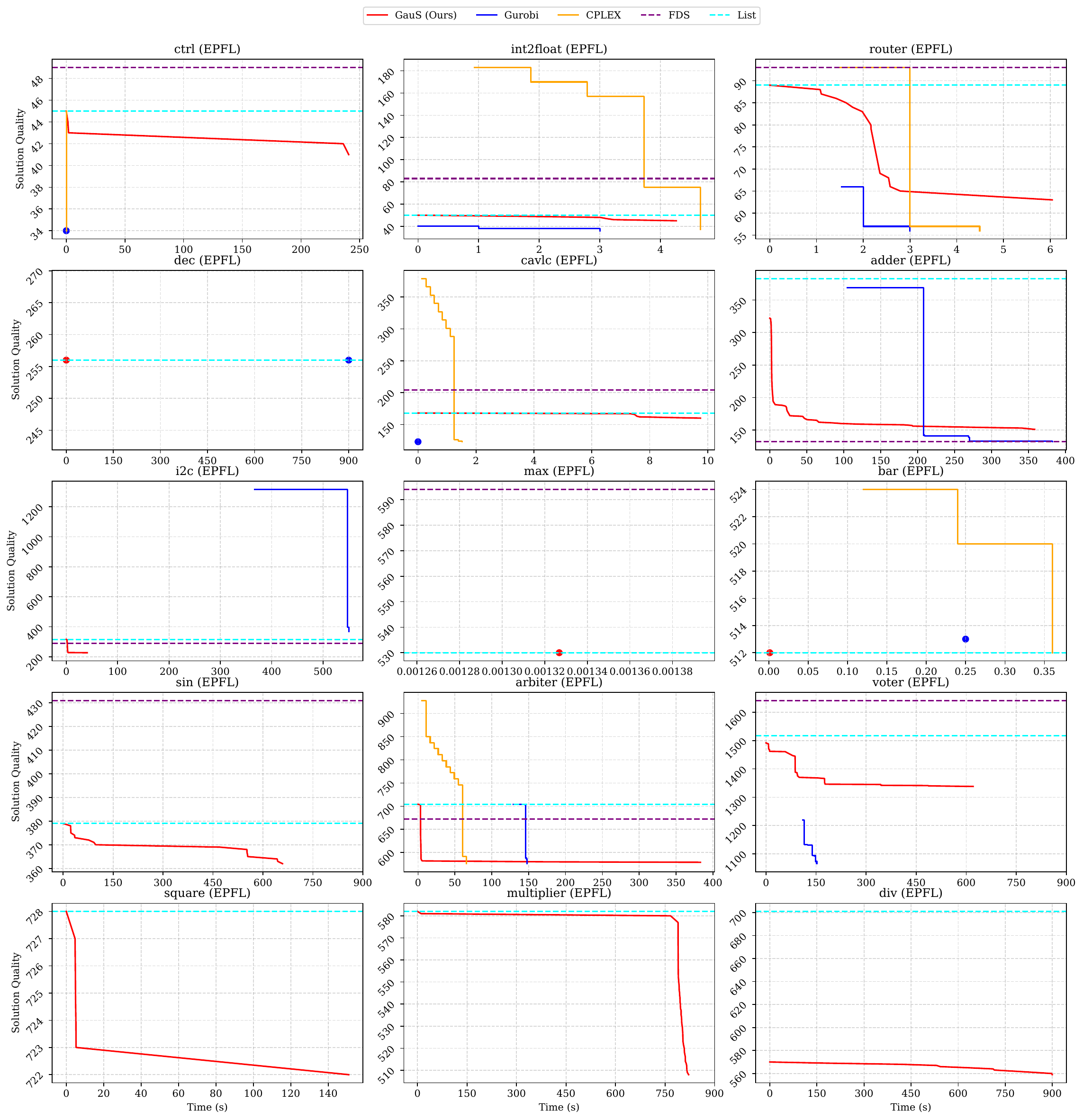}
    \caption{Comprehensive tradeoff comparison of EPFL on \hyperlink{formB}{\textbf{Formulation B: }} \textit{Latency constrained, memory footprint optimization} --- 
    \small{A 15-minute time limit is applied to all methods. 
    Heuristics such as FDS and List Scheduling are represented as horizontal lines to reflect their fast convergence to final solution quality.}}
    \label{fig:formB_epfl}
\end{figure*}

%% file: figures/tradeoff_rw_formC.tex
\begin{figure*}[!ht]
    \centering
    \includegraphics[width=\textwidth]{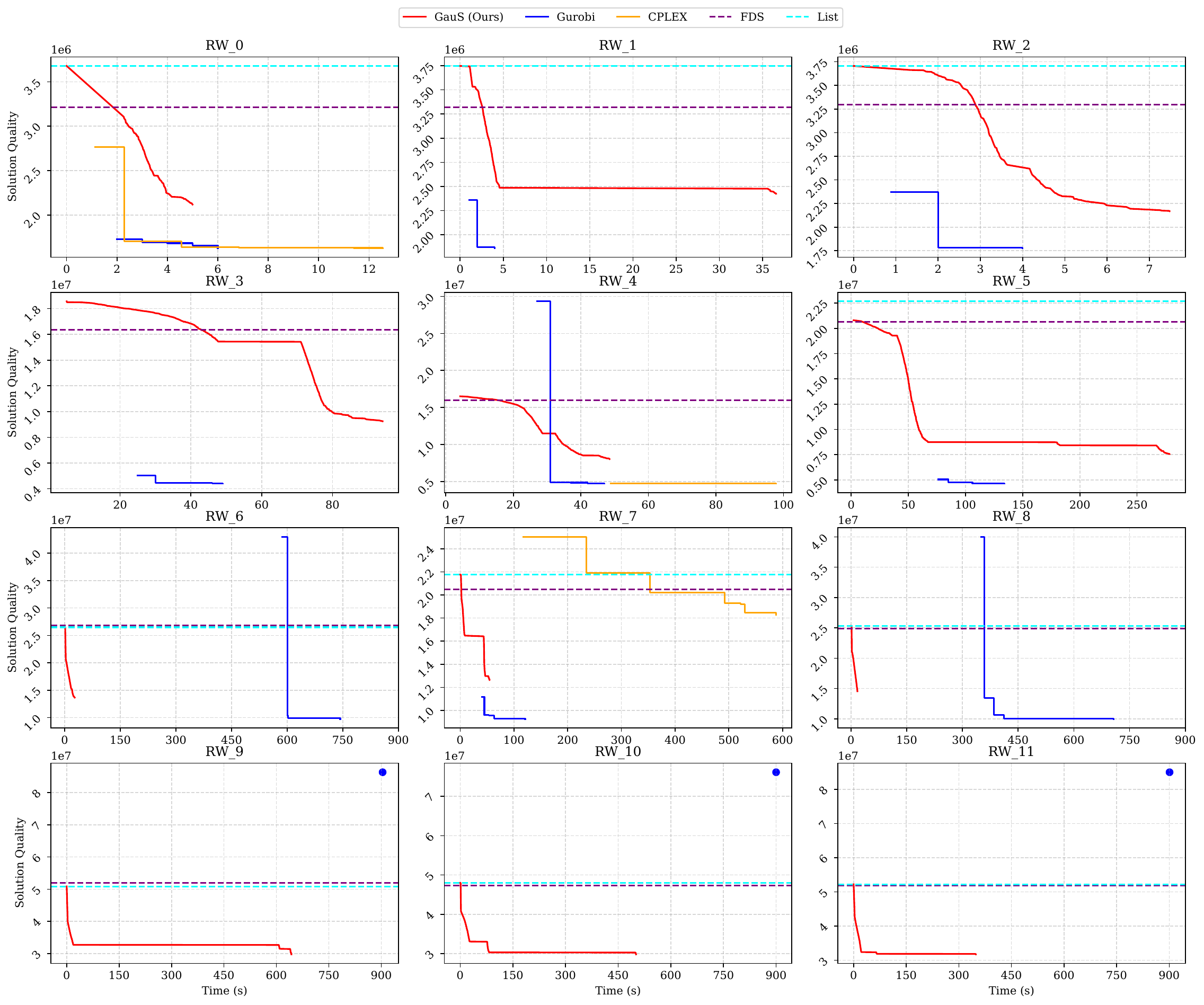}
    \caption{Comprehensive tradeoff comparison of RW on \hyperlink{formC}{\textbf{Formulation C: }} \textit{Latency, resource, and recurrences constrained, memory footprint modulo optimization} --- 
    \small{A 15-minute time limit is applied to all methods.
    Heuristics such as FDS and List Scheduling are represented as horizontal lines to reflect their fast convergence to final solution quality.
    }}
    \label{fig:formC_rw}
\end{figure*}

%% file: figures/tradeoff_epfl_formC.tex
\begin{figure*}[htbp]
    \centering
    \includegraphics[width=\textwidth]{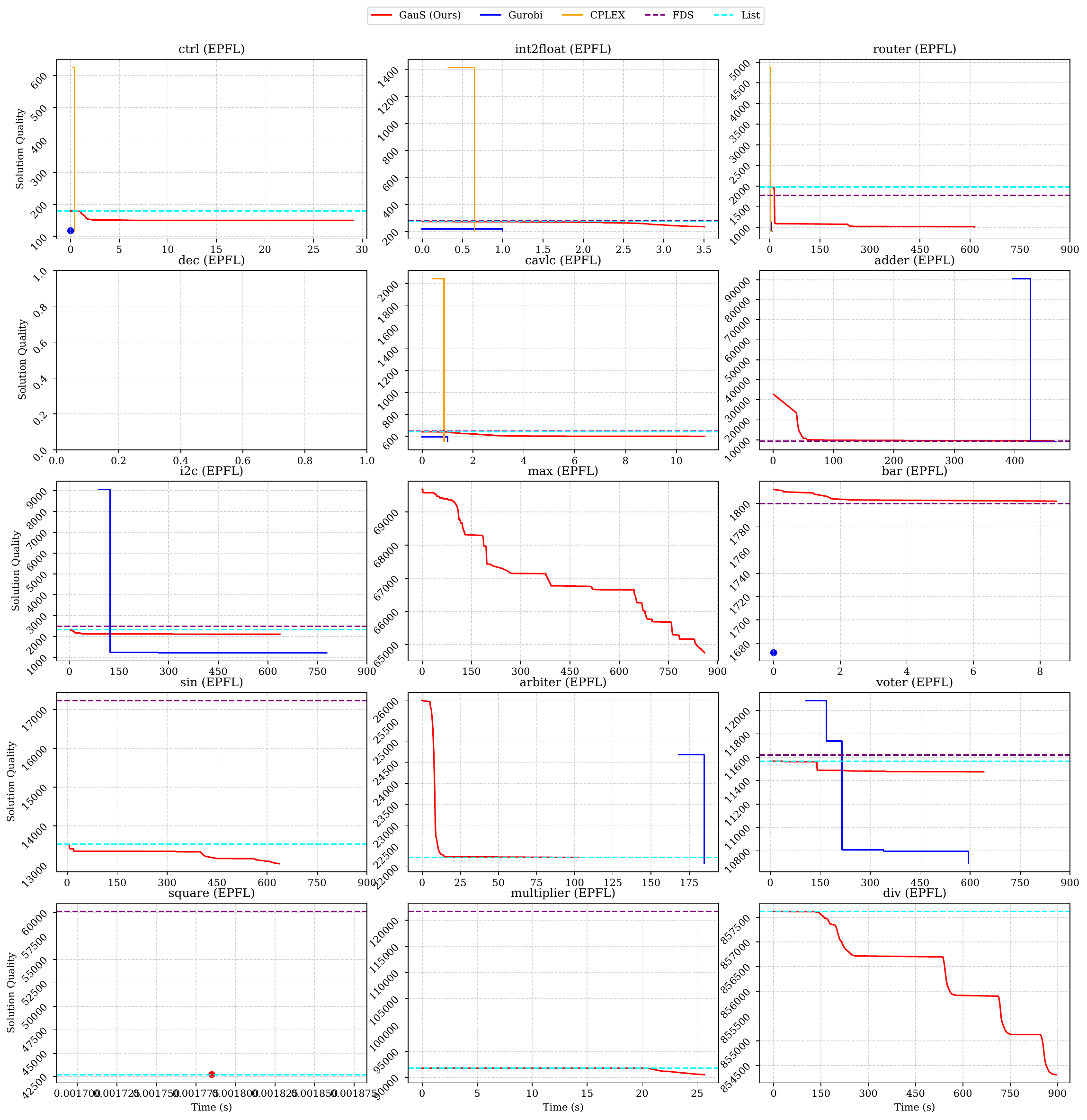}
    \caption{Comprehensive tradeoff comparison of EPFL on \hyperlink{formC}{\textbf{Formulation C: }} \textit{Latency, resource, and recurrences constrained, memory footprint modulo optimization} --- 
    \small{A 15-minute time limit is applied to all methods. 
    Heuristics such as FDS and List Scheduling are represented as horizontal lines to reflect their fast convergence to final solution quality.
    There is no feasible solution for dec, thus plot is left empty. }}
    \label{fig:formC_epfl}
\end{figure*}

%% file: checklist.tex
\input{a0-impact}
\section*{NeurIPS Paper Checklist}

\begin{enumerate}

\item {\bf Claims}
    \item[] Question: Do the main claims made in the abstract and introduction accurately reflect the paper's contributions and scope?
    \item[] Answer: \answerYes{} 
    \item[] Justification: The main claims reflect the papers's contribution and scope and are supported emperically.
    \item[] Guidelines:
    \begin{itemize}
        \item The answer \answerNA{} means that the abstract and introduction do not include the claims made in the paper.
        \item The abstract and/or introduction should clearly state the claims made, including the contributions made in the paper and important assumptions and limitations. A \answerNo{} or \answerNA{} answer to this question will not be perceived well by the reviewers. 
        \item The claims made should match theoretical and experimental results, and reflect how much the results can be expected to generalize to other settings. 
        \item It is fine to include aspirational goals as motivation as long as it is clear that these goals are not attained by the paper. 
    \end{itemize}

\item {\bf Limitations}
    \item[] Question: Does the paper discuss the limitations of the work performed by the authors?
    \item[] Answer: \answerYes{} 
    \item[] Justification: The limitations are discussed in the conclusion and the appendices.
    \item[] Guidelines:
    \begin{itemize}
        \item The answer \answerNA{} means that the paper has no limitation while the answer \answerNo{} means that the paper has limitations, but those are not discussed in the paper. 
        \item The authors are encouraged to create a separate ``Limitations'' section in their paper.
        \item The paper should point out any strong assumptions and how robust the results are to violations of these assumptions (e.g., independence assumptions, noiseless settings, model well-specification, asymptotic approximations only holding locally). The authors should reflect on how these assumptions might be violated in practice and what the implications would be.
        \item The authors should reflect on the scope of the claims made, e.g., if the approach was only tested on a few datasets or with a few runs. In general, empirical results often depend on implicit assumptions, which should be articulated.
        \item The authors should reflect on the factors that influence the performance of the approach. For example, a facial recognition algorithm may perform poorly when image resolution is low or images are taken in low lighting. Or a speech-to-text system might not be used reliably to provide closed captions for online lectures because it fails to handle technical jargon.
        \item The authors should discuss the computational efficiency of the proposed algorithms and how they scale with dataset size.
        \item If applicable, the authors should discuss possible limitations of their approach to address problems of privacy and fairness.
        \item While the authors might fear that complete honesty about limitations might be used by reviewers as grounds for rejection, a worse outcome might be that reviewers discover limitations that aren't acknowledged in the paper. The authors should use their best judgment and recognize that individual actions in favor of transparency play an important role in developing norms that preserve the integrity of the community. Reviewers will be specifically instructed to not penalize honesty concerning limitations.
    \end{itemize}

\item {\bf Theory assumptions and proofs}
    \item[] Question: For each theoretical result, does the paper provide the full set of assumptions and a complete (and correct) proof?
    \item[] Answer: \answerNA{} 
    \item[] Justification: This paper does not include any theoretical results. 
    \item[] Guidelines:
    \begin{itemize}
        \item The answer \answerNA{} means that the paper does not include theoretical results. 
        \item All the theorems, formulas, and proofs in the paper should be numbered and cross-referenced.
        \item All assumptions should be clearly stated or referenced in the statement of any theorems.
        \item The proofs can either appear in the main paper or the supplemental material, but if they appear in the supplemental material, the authors are encouraged to provide a short proof sketch to provide intuition. 
        \item Inversely, any informal proof provided in the core of the paper should be complemented by formal proofs provided in appendix or supplemental material.
        \item Theorems and Lemmas that the proof relies upon should be properly referenced. 
    \end{itemize}

    \item {\bf Experimental result reproducibility}
    \item[] Question: Does the paper fully disclose all the information needed to reproduce the main experimental results of the paper to the extent that it affects the main claims and/or conclusions of the paper (regardless of whether the code and data are provided or not)?
    \item[] Answer: \answerYes{} 
    \item[] Justification: All the experiment details are included in the evaluation and appendices. We also empirically show that the experimental results are robust to parameter selection in general.
    \item[] Guidelines:
    \begin{itemize}
        \item The answer \answerNA{} means that the paper does not include experiments.
        \item If the paper includes experiments, a \answerNo{} answer to this question will not be perceived well by the reviewers: Making the paper reproducible is important, regardless of whether the code and data are provided or not.
        \item If the contribution is a dataset and\slash or model, the authors should describe the steps taken to make their results reproducible or verifiable. 
        \item Depending on the contribution, reproducibility can be accomplished in various ways. For example, if the contribution is a novel architecture, describing the architecture fully might suffice, or if the contribution is a specific model and empirical evaluation, it may be necessary to either make it possible for others to replicate the model with the same dataset, or provide access to the model. In general. releasing code and data is often one good way to accomplish this, but reproducibility can also be provided via detailed instructions for how to replicate the results, access to a hosted model (e.g., in the case of a large language model), releasing of a model checkpoint, or other means that are appropriate to the research performed.
        \item While NeurIPS does not require releasing code, the conference does require all submissions to provide some reasonable avenue for reproducibility, which may depend on the nature of the contribution. For example
        \begin{enumerate}
            \item If the contribution is primarily a new algorithm, the paper should make it clear how to reproduce that algorithm.
            \item If the contribution is primarily a new model architecture, the paper should describe the architecture clearly and fully.
            \item If the contribution is a new model (e.g., a large language model), then there should either be a way to access this model for reproducing the results or a way to reproduce the model (e.g., with an open-source dataset or instructions for how to construct the dataset).
            \item We recognize that reproducibility may be tricky in some cases, in which case authors are welcome to describe the particular way they provide for reproducibility. In the case of closed-source models, it may be that access to the model is limited in some way (e.g., to registered users), but it should be possible for other researchers to have some path to reproducing or verifying the results.
        \end{enumerate}
    \end{itemize}

\item {\bf Open access to data and code}
    \item[] Question: Does the paper provide open access to the data and code, with sufficient instructions to faithfully reproduce the main experimental results, as described in supplemental material?
    \item[] Answer: \answerYes{} 
    \item[] Justification: The code and data are submitted for experimental results reproduction.
    \item[] Guidelines:
    \begin{itemize}
        \item The answer \answerNA{} means that paper does not include experiments requiring code.
        \item Please see the NeurIPS code and data submission guidelines (\url{https://neurips.cc/public/guides/CodeSubmissionPolicy}) for more details.
        \item While we encourage the release of code and data, we understand that this might not be possible, so \answerNo{} is an acceptable answer. Papers cannot be rejected simply for not including code, unless this is central to the contribution (e.g., for a new open-source benchmark).
        \item The instructions should contain the exact command and environment needed to run to reproduce the results. See the NeurIPS code and data submission guidelines (\url{https://neurips.cc/public/guides/CodeSubmissionPolicy}) for more details.
        \item The authors should provide instructions on data access and preparation, including how to access the raw data, preprocessed data, intermediate data, and generated data, etc.
        \item The authors should provide scripts to reproduce all experimental results for the new proposed method and baselines. If only a subset of experiments are reproducible, they should state which ones are omitted from the script and why.
        \item At submission time, to preserve anonymity, the authors should release anonymized versions (if applicable).
        \item Providing as much information as possible in supplemental material (appended to the paper) is recommended, but including URLs to data and code is permitted.
    \end{itemize}

\item {\bf Experimental setting/details}
    \item[] Question: Does the paper specify all the training and test details (e.g., data splits, hyperparameters, how they were chosen, type of optimizer) necessary to understand the results?
    \item[] Answer: \answerYes{} 
    \item[] Justification: Details can be found in evaluation and appendices~\ref{app:hp}.
    \item[] Guidelines:
    \begin{itemize}
        \item The answer \answerNA{} means that the paper does not include experiments.
        \item The experimental setting should be presented in the core of the paper to a level of detail that is necessary to appreciate the results and make sense of them.
        \item The full details can be provided either with the code, in appendix, or as supplemental material.
    \end{itemize}

\item {\bf Experiment statistical significance}
    \item[] Question: Does the paper report error bars suitably and correctly defined or other appropriate information about the statistical significance of the experiments?
    \item[] Answer: \answerYes{} 
    \item[] Justification: The proposed method is deterministic without variance, thus we do not have error bars. We extensively evaluate it on various benchmarks to demonstrate significance.
    \item[] Guidelines:
    \begin{itemize}
        \item The answer \answerNA{} means that the paper does not include experiments.
        \item The authors should answer \answerYes{} if the results are accompanied by error bars, confidence intervals, or statistical significance tests, at least for the experiments that support the main claims of the paper.
        \item The factors of variability that the error bars are capturing should be clearly stated (for example, train/test split, initialization, random drawing of some parameter, or overall run with given experimental conditions).
        \item The method for calculating the error bars should be explained (closed form formula, call to a library function, bootstrap, etc.)
        \item The assumptions made should be given (e.g., Normally distributed errors).
        \item It should be clear whether the error bar is the standard deviation or the standard error of the mean.
        \item It is OK to report 1-sigma error bars, but one should state it. The authors should preferably report a 2-sigma error bar than state that they have a 96\% CI, if the hypothesis of Normality of errors is not verified.
        \item For asymmetric distributions, the authors should be careful not to show in tables or figures symmetric error bars that would yield results that are out of range (e.g., negative error rates).
        \item If error bars are reported in tables or plots, the authors should explain in the text how they were calculated and reference the corresponding figures or tables in the text.
    \end{itemize}

\item {\bf Experiments compute resources}
    \item[] Question: For each experiment, does the paper provide sufficient information on the computer resources (type of compute workers, memory, time of execution) needed to reproduce the experiments?
    \item[] Answer: \answerYes{} 
    \item[] Justification: Evaluation, Appendix~\ref{app:hp}, and Appendix~\ref{app:tradeoff} provide all the details.
    \item[] Guidelines:
    \begin{itemize}
        \item The answer \answerNA{} means that the paper does not include experiments.
        \item The paper should indicate the type of compute workers CPU or GPU, internal cluster, or cloud provider, including relevant memory and storage.
        \item The paper should provide the amount of compute required for each of the individual experimental runs as well as estimate the total compute. 
        \item The paper should disclose whether the full research project required more compute than the experiments reported in the paper (e.g., preliminary or failed experiments that didn't make it into the paper). 
    \end{itemize}
    
\item {\bf Code of ethics}
    \item[] Question: Does the research conducted in the paper conform, in every respect, with the NeurIPS Code of Ethics \url{https://neurips.cc/public/EthicsGuidelines}?
    \item[] Answer: \answerYes{} 
    \item[] Justification: We reviewed the NeurIPS code of Ethics and conform with it.
    \item[] Guidelines:
    \begin{itemize}
        \item The answer \answerNA{} means that the authors have not reviewed the NeurIPS Code of Ethics.
        \item If the authors answer \answerNo, they should explain the special circumstances that require a deviation from the Code of Ethics.
        \item The authors should make sure to preserve anonymity (e.g., if there is a special consideration due to laws or regulations in their jurisdiction).
    \end{itemize}

\item {\bf Broader impacts}
    \item[] Question: Does the paper discuss both potential positive societal impacts and negative societal impacts of the work performed?
    \item[] Answer: \answerNA{} 
    \item[] Justification: Our method solely focuses on improving the scalability and efficiency of the scheduling algorithm. It does not impact the society at large.
    \item[] Guidelines:
    \begin{itemize}
        \item The answer \answerNA{} means that there is no societal impact of the work performed.
        \item If the authors answer \answerNA{} or \answerNo, they should explain why their work has no societal impact or why the paper does not address societal impact.
        \item Examples of negative societal impacts include potential malicious or unintended uses (e.g., disinformation, generating fake profiles, surveillance), fairness considerations (e.g., deployment of technologies that could make decisions that unfairly impact specific groups), privacy considerations, and security considerations.
        \item The conference expects that many papers will be foundational research and not tied to particular applications, let alone deployments. However, if there is a direct path to any negative applications, the authors should point it out. For example, it is legitimate to point out that an improvement in the quality of generative models could be used to generate Deepfakes for disinformation. On the other hand, it is not needed to point out that a generic algorithm for optimizing neural networks could enable people to train models that generate Deepfakes faster.
        \item The authors should consider possible harms that could arise when the technology is being used as intended and functioning correctly, harms that could arise when the technology is being used as intended but gives incorrect results, and harms following from (intentional or unintentional) misuse of the technology.
        \item If there are negative societal impacts, the authors could also discuss possible mitigation strategies (e.g., gated release of models, providing defenses in addition to attacks, mechanisms for monitoring misuse, mechanisms to monitor how a system learns from feedback over time, improving the efficiency and accessibility of ML).
    \end{itemize}
    
\item {\bf Safeguards}
    \item[] Question: Does the paper describe safeguards that have been put in place for responsible release of data or models that have a high risk for misuse (e.g., pre-trained language models, image generators, or scraped datasets)?
    \item[] Answer: \answerNA{} 
    \item[] Justification: We do not add any realistic new data or models. 
    \item[] Guidelines:
    \begin{itemize}
        \item The answer \answerNA{} means that the paper poses no such risks.
        \item Released models that have a high risk for misuse or dual-use should be released with necessary safeguards to allow for controlled use of the model, for example by requiring that users adhere to usage guidelines or restrictions to access the model or implementing safety filters. 
        \item Datasets that have been scraped from the Internet could pose safety risks. The authors should describe how they avoided releasing unsafe images.
        \item We recognize that providing effective safeguards is challenging, and many papers do not require this, but we encourage authors to take this into account and make a best faith effort.
    \end{itemize}

\item {\bf Licenses for existing assets}
    \item[] Question: Are the creators or original owners of assets (e.g., code, data, models), used in the paper, properly credited and are the license and terms of use explicitly mentioned and properly respected?
    \item[] Answer: \answerYes{} 
    \item[] Justification: Yes, we comply with the licenses and terms of use as specified by the original creators of the data and code involved in this paper.
    \item[] Guidelines:
    \begin{itemize}
        \item The answer \answerNA{} means that the paper does not use existing assets.
        \item The authors should cite the original paper that produced the code package or dataset.
        \item The authors should state which version of the asset is used and, if possible, include a URL.
        \item The name of the license (e.g., CC-BY 4.0) should be included for each asset.
        \item For scraped data from a particular source (e.g., website), the copyright and terms of service of that source should be provided.
        \item If assets are released, the license, copyright information, and terms of use in the package should be provided. For popular datasets, \url{paperswithcode.com/datasets} has curated licenses for some datasets. Their licensing guide can help determine the license of a dataset.
        \item For existing datasets that are re-packaged, both the original license and the license of the derived asset (if it has changed) should be provided.
        \item If this information is not available online, the authors are encouraged to reach out to the asset's creators.
    \end{itemize}

\item {\bf New assets}
    \item[] Question: Are new assets introduced in the paper well documented and is the documentation provided alongside the assets?
    \item[] Answer: \answerYes{} 
    \item[] Justification: We introduces new synthetic data. The data generation code is provided and the format is documented. No privacy or copyright issue is involved as the new data is purely synthetic. 
    \item[] Guidelines:
    \begin{itemize}
        \item The answer \answerNA{} means that the paper does not release new assets.
        \item Researchers should communicate the details of the dataset\slash code\slash model as part of their submissions via structured templates. This includes details about training, license, limitations, etc. 
        \item The paper should discuss whether and how consent was obtained from people whose asset is used.
        \item At submission time, remember to anonymize your assets (if applicable). You can either create an anonymized URL or include an anonymized zip file.
    \end{itemize}

\item {\bf Crowdsourcing and research with human subjects}
    \item[] Question: For crowdsourcing experiments and research with human subjects, does the paper include the full text of instructions given to participants and screenshots, if applicable, as well as details about compensation (if any)? 
    \item[] Answer: \answerNA{} 
    \item[] Justification: This paper does not involve crowdsourcing not research with human subjects. 
    \item[] Guidelines:
    \begin{itemize}
        \item The answer \answerNA{} means that the paper does not involve crowdsourcing nor research with human subjects.
        \item Including this information in the supplemental material is fine, but if the main contribution of the paper involves human subjects, then as much detail as possible should be included in the main paper. 
        \item According to the NeurIPS Code of Ethics, workers involved in data collection, curation, or other labor should be paid at least the minimum wage in the country of the data collector. 
    \end{itemize}

\item {\bf Institutional review board (IRB) approvals or equivalent for research with human subjects}
    \item[] Question: Does the paper describe potential risks incurred by study participants, whether such risks were disclosed to the subjects, and whether Institutional Review Board (IRB) approvals (or an equivalent approval/review based on the requirements of your country or institution) were obtained?
    \item[] Answer: \answerNA{} 
    \item[] Justification: This paper does not involve crowdsourcing not research with human subjects. 
    \item[] Guidelines:
    \begin{itemize}
        \item The answer \answerNA{} means that the paper does not involve crowdsourcing nor research with human subjects.
        \item Depending on the country in which research is conducted, IRB approval (or equivalent) may be required for any human subjects research. If you obtained IRB approval, you should clearly state this in the paper. 
        \item We recognize that the procedures for this may vary significantly between institutions and locations, and we expect authors to adhere to the NeurIPS Code of Ethics and the guidelines for their institution. 
        \item For initial submissions, do not include any information that would break anonymity (if applicable), such as the institution conducting the review.
    \end{itemize}

\item {\bf Declaration of LLM usage}
    \item[] Question: Does the paper describe the usage of LLMs if it is an important, original, or non-standard component of the core methods in this research? Note that if the LLM is used only for writing, editing, or formatting purposes and does \emph{not} impact the core methodology, scientific rigor, or originality of the research, declaration is not required.
    \item[] Answer: \answerNA{} 
    \item[] Justification: We use LLM primarily for editing purpose and basic code assistance.
    \item[] Guidelines:
    \begin{itemize}
        \item The answer \answerNA{} means that the core method development in this research does not involve LLMs as any important, original, or non-standard components.
        \item Please refer to our LLM policy in the NeurIPS handbook for what should or should not be described.
    \end{itemize}

\end{enumerate}

%% file: a0-impact.tex

%% file: references.bib
@string{ISCA      = {Int'l Symp. on Computer Architecture (ISCA)}}

@string{MICRO     = {Int'l Symp. on Microarchitecture (MICRO)}}

@string{DAC      = {Design Automation Conf. (DAC)}}

@string{ICCAD    = {Int'l Conf. on Computer-Aided Design (ICCAD)}}

@inproceedings{liu2024differentiable,
  title={Differentiable combinatorial scheduling at scale},
  author={Liu, Mingju and Li, Yingjie and Yin, Jiaqi and Zhang, Zhiru and Yu, Cunxi},
  booktitle={Proceedings of the 41st International Conference on Machine Learning},
  pages={31464--31476},
  year={2024}
}

@article{rau1981some,
  title={Some scheduling techniques and an easily schedulable horizontal architecture for high performance scientific computing},
  author={Rau, B Ramakrishna and Glaeser, Christopher D},
  journal={ACM SIGMICRO Newsletter},
  volume={12},
  number={4},
  pages={183--198},
  year={1981},
  publisher={ACM New York, NY, USA}
}

@inproceedings{lam1988software,
  title={Software pipelining: An effective scheduling technique for VLIW machines},
  author={Lam, Monica},
  booktitle={Proceedings of the ACM SIGPLAN 1988 conference on Programming Language design and Implementation},
  pages={318--328},
  year={1988}
}

@article{amaru2015epfl,
  title={The EPFL combinational benchmark suite},
  author={Amar{\'u}, Luca and Gaillardon, Pierre-Emmanuel and De Micheli, Giovanni},
  journal={Hypotenuse},
  volume={256},
  number={128},
  pages={214335},
  year={2015}
}

@inproceedings{zhang2013sdc,
  title={SDC-based modulo scheduling for pipeline synthesis},
  author={Zhang, Zhiru and Liu, Bin},
  booktitle={2013 IEEE/ACM International Conference on Computer-Aided Design (ICCAD)},
  pages={211--218},
  year={2013},
  organization={IEEE}
}

@article{oppermann2019exact,
  title={Exact and practical modulo scheduling for high-level synthesis},
  author={Oppermann, Julian and Reuter-Oppermann, Melanie and Sommer, Lukas and Koch, Andreas and Sinnen, Oliver},
  journal={ACM Transactions on Reconfigurable Technology and Systems (TRETS)},
  volume={12},
  number={2},
  pages={1--26},
  year={2019},
  publisher={ACM New York, NY, USA}
}

@article{soi2025optimal,
  title={Optimal Software Pipelining and Warp Specialization for Tensor Core GPUs},
  author={Soi, Rupanshu and Yadav, Rohan and Kjolstad, Fredrik and Aiken, Alex and Dehnavi, Maryam Mehri and Garland, Michael and Bauer, Michael},
  journal={arXiv preprint arXiv:2512.18134},
  year={2025}
}

@inproceedings{cong2006efficient,
  title={An efficient and versatile scheduling algorithm based on SDC formulation},
  author={Cong, Jason and Zhang, Zhiru},
  booktitle={Proceedings of the 43rd annual Design Automation Conference},
  pages={433--438},
  year={2006}
}

@article{cplex,
  title={Solving mixed-integer quadratic programming problems with IBM-CPLEX: a progress report},
  author={Bliek1{\'u}, Christian and Bonami, Pierre and Lodi, Andrea},
  journal={Proceedings of the twenty-sixth RAMP symposium},
  pages={16--17},
  year={2014}
}

@misc{gurobi,
  author = {{Gurobi Optimization, LLC}},
  title = {{Gurobi Optimizer Reference Manual}},
  year = 2023,
  url = "https://www.gurobi.com"
}

@inproceedings{kondratyev2011realistic,
  title={Realistic performance-constrained pipelining in high-level synthesis},
  author={Kondratyev, Alex and Lavagno, Luciano and Meyer, Mike and Watanabe, Yosinori},
  booktitle={2011 Design, Automation \& Test in Europe},
  pages={1--6},
  year={2011},
  organization={IEEE}
}

@inproceedings{parker1986maha,
  title={MAHA: A program for datapath synthesis},
  author={Parker, Alice C and Pizarro, Jorge and Mlinar, Mitch},
  booktitle={23rd ACM/IEEE Design Automation Conference},
  pages={461--466},
  year={1986},
  organization={IEEE}
}

@inproceedings{paulin1987force,
author = {Paulin, P. G. and Knight, J. P.},
title = {Force-directed scheduling in automatic data path synthesis},
year = {1987},
isbn = {0818607815},
publisher = {Association for Computing Machinery},
address = {New York, NY, USA},
url = {https://doi.org/10.1145/37888.37918},
doi = {10.1145/37888.37918},
booktitle = {Proceedings of the 24th ACM/IEEE Design Automation Conference},
pages = {195–202},
numpages = {8},
location = {Miami Beach, Florida, USA},
series = {DAC '87}
}

@INPROCEEDINGS{verhaegh1992force,
  author={Verhaegh and Lippens and Aarts and Korst and van der Werf and van Meerbergen},
  booktitle={1992 IEEE/ACM International Conference on Computer-Aided Design}, 
  title={Efficiency improvements for force-directed scheduling}, 
  year={1992},
  volume={},
  number={},
  pages={286-291},
  doi={10.1109/ICCAD.1992.279359}}

@article{ku2002relative,
  title={Relative scheduling under timing constraints: Algorithms for high-level synthesis of digital circuits},
  author={Ku, David C and De Mitcheli, G},
  journal={IEEE Transactions on Computer-Aided Design of Integrated Circuits and Systems},
  volume={11},
  number={6},
  pages={696--718},
  year={2002},
  publisher={IEEE}
}

@article{radivojevic1996new,
  title={A new symbolic technique for control-dependent scheduling},
  author={Radivojevic, Ivan and Brewer, Forrest},
  journal={IEEE transactions on computer-aided design of integrated circuits and systems},
  volume={15},
  number={1},
  pages={45--57},
  year={1996},
  publisher={IEEE}
}

@inproceedings{fan2005cost,
  title={Cost sensitive modulo scheduling in a loop accelerator synthesis system},
  author={Fan, Kevin and Kudlur, Manjunath and Park, Hyunchul and Mahlke, Scott},
  booktitle={38th Annual IEEE/ACM International Symposium on Microarchitecture (MICRO'05)},
  pages={12--pp},
  year={2005},
  organization={IEEE}
}

@inproceedings{fan2008modulo,
  title={Modulo scheduling for highly customized datapaths to increase hardware reusability},
  author={Fan, Kevin and Park, Hyun hul and Kudlur, Manjunath and Mahlke, S ott},
  booktitle={Proceedings of the 6th annual IEEE/ACM international symposium on Code generation and optimization},
  pages={124--133},
  year={2008}
}

@inproceedings{cai2025smoothe,
  title={Smoothe: Differentiable e-graph extraction},
  author={Cai, Yaohui and Yang, Kaixin and Deng, Chenhui and Yu, Cunxi and Zhang, Zhiru},
  booktitle={Proceedings of the 30th ACM International Conference on Architectural Support for Programming Languages and Operating Systems, Volume 1},
  pages={1020--1034},
  year={2025}
}

@inproceedings{wu2023gamora,
  title={Gamora: Graph learning based symbolic reasoning for large-scale boolean networks},
  author={Wu, Nan and Li, Yingjie and Hao, Cong and Dai, Steve and Yu, Cunxi and Xie, Yuan},
  booktitle={2023 60th ACM/IEEE Design Automation Conference (DAC)},
  pages={1--6},
  year={2023},
  organization={IEEE}
}

@inproceedings{lin2019dreamplace,
  title={Dreamplace: Deep learning toolkit-enabled gpu acceleration for modern vlsi placement},
  author={Lin, Yibo and Dhar, Shounak and Li, Wuxi and Ren, Haoxing and Khailany, Brucek and Pan, David Z},
  booktitle={Proceedings of the 56th Annual Design Automation Conference 2019},
  pages={1--6},
  year={2019}
}

@article{bengio2021machine,
  title={Machine learning for combinatorial optimization: a methodological tour d’horizon},
  author={Bengio, Yoshua and Lodi, Andrea and Prouvost, Antoine},
  journal={European Journal of Operational Research},
  volume={290},
  number={2},
  pages={405--421},
  year={2021},
  publisher={Elsevier}
}

@inproceedings{zhang2022gatspi,
  title={GATSPI: GPU accelerated gate-level simulation for power improvement},
  author={Zhang, Yanqing and Ren, Haoxing and Sridharan, Akshay and Khailany, Brucek},
  booktitle={Proceedings of the 59th ACM/IEEE Design Automation Conference},
  pages={1231--1236},
  year={2022}
}

@article{hwang2002formal,
  title={A formal approach to the scheduling problem in high level synthesis},
  author={Hwang, C-T and Lee, J-H and Hsu, Y-C},
  journal={IEEE Transactions on Computer-Aided Design of Integrated Circuits and Systems},
  volume={10},
  number={4},
  pages={464--475},
  year={2002},
  publisher={IEEE}
}

@ARTICLE{gebots1993cost,
  author={Gebotys, C.H. and Elmasry, M.I.},
  journal={IEEE Transactions on Computer-Aided Design of Integrated Circuits and Systems}, 
  title={Global optimization approach for architectural synthesis}, 
  year={1993},
  volume={12},
  number={9},
  pages={1266-1278},
  doi={10.1109/43.240074}}

@inproceedings{zhang2004sat,
  title={A SAT Based Scheduler for Tournament Schedules.},
  author={Zhang, Hantao and Li, Dapeng and Shen, Haiou},
  booktitle={SAT},
  year={2004}
}

@inproceedings{chen2024syn,
  title={E-syn: E-graph rewriting with technology-aware cost functions for logic synthesis},
  author={Chen, Chen and Hu, Guangyu and Zuo, Dongsheng and Yu, Cunxi and Ma, Yuzhe and Zhang, Hongce},
  booktitle={Proceedings of the 61st ACM/IEEE Design Automation Conference},
  pages={1--6},
  year={2024}
}

@article{jang2016categorical,
  title={Categorical reparameterization with gumbel-softmax},
  author={Jang, Eric and Gu, Shixiang and Poole, Ben},
  journal={arXiv preprint arXiv:1611.01144},
  year={2016}
}

@article{powell1969method,
  title={A method for nonlinear constraints in minimization problems},
  author={Powell, Michael JD},
  journal={Optimization},
  pages={283--298},
  year={1969},
  publisher={Academic Press}
}

@article{hestenes1969multiplier,
  title={Multiplier and gradient methods},
  author={Hestenes, Magnus R},
  journal={Journal of optimization theory and applications},
  volume={4},
  number={5},
  pages={303--320},
  year={1969},
  publisher={Springer}
}

@inproceedings{abts2020think,
  title={Think fast: A tensor streaming processor (TSP) for accelerating deep learning workloads},
  author={Abts, Dennis and Ross, Jonathan and Sparling, Jonathan and Wong-VanHaren, Mark and Baker, Max and Hawkins, Tom and Bell, Andrew and Thompson, John and Kahsai, Temesghen and Kimmell, Garrin and others},
  booktitle={2020 ACM/IEEE 47th Annual International Symposium on Computer Architecture (ISCA)},
  pages={145--158},
  year={2020},
  organization={IEEE}
}

@inproceedings{zhang2025cypress,
  title={Cypress: VLSI-Inspired PCB Placement with GPU Acceleration},
  author={Zhang, Niansong and Agnesina, Anthony and Shbat, Noor and Leader, Yuval and Zhang, Zhiru and Ren, Haoxing},
  booktitle={Proceedings of the 2025 International Symposium on Physical Design},
  pages={31--41},
  year={2025}
}

@article{guo2025pimsynth,
  title={PIMsynth: A Unified Compiler Framework for Bit-Serial Processing-In-Memory Architectures},
  author={Guo, Deyuan and Gholamrezaei, Mohammadhosein and Hofmann, Matthew and Venkat, Ashish and Zhang, Zhiru and Skadron, Kevin},
  journal={IEEE Computer Architecture Letters},
  year={2025},
  publisher={IEEE}
}

@inproceedings{seshadri2017ambit,
  title={Ambit: In-memory accelerator for bulk bitwise operations using commodity DRAM technology},
  author={Seshadri, Vivek and Lee, Donghyuk and Mullins, Thomas and Hassan, Hasan and Boroumand, Amirali and Kim, Jeremie and Kozuch, Michael A and Mutlu, Onur and Gibbons, Phillip B and Mowry, Todd C},
  booktitle={Proceedings of the 50th Annual IEEE/ACM International Symposium on Microarchitecture},
  pages={273--287},
  year={2017}
}

@article{leiserson1991retiming,
  title={Retiming synchronous circuitry},
  author={Leiserson, Charles E and Saxe, James B},
  journal={Algorithmica},
  volume={6},
  number={1},
  pages={5--35},
  year={1991},
  publisher={Springer}
}
